\documentclass[sigconf]{acmart}
\AtBeginDocument{%
  }

\usepackage{multirow}
\usepackage{colortbl}
\usepackage{subfigure}
\usepackage{diagbox}
\usepackage{color}
\usepackage{rotating}
\usepackage{hyperref}
\usepackage{balance}
\usepackage{amsmath}
\usepackage[normalem]{ulem}
\useunder{\uline}{\ul}{}
\usepackage{tcolorbox}
\newtcolorbox{examplebox}{
  colback=gray!10,
  colframe=gray!50,
  arc=4pt,
  boxrule=0.4pt,
  left=6pt,
  right=6pt,
  top=4pt,
  bottom=4pt,
  fontupper=\footnotesize\ttfamily  
}

\setcopyright{acmlicensed}
\copyrightyear{2018}
\acmYear{2018}
\acmDOI{XXXXXXX.XXXXXXX}
\acmConference[Conference acronym 'XX]{Make sure to enter the correct
  conference title from your rights confirmation email}{June 03--05,
  2018}{Woodstock, NY}
\acmISBN{978-1-4503-XXXX-X/2018/06}

\begin{document}

\title{Time Imprint: Learning Time-Aware Representations in Multi-Modal Knowledge Graphs}

\author{Pengyu Zhang}
\email{p.zhang@uva.nl}
\orcid{0000-0001-5111-4487}
\affiliation{
  \institution{University of Amsterdam}
  \city{Amsterdam}
  \country{The Netherlands}
}

\author{Klim Zaporojets}
\orcid{0000-0003-4988-978X}
\affiliation{
  \institution{Aarhus University}
  \city{Aarhus}
  \country{Denmark}
}

\author{Congfeng Cao}
\orcid{0000-0001-9011-3807}
\affiliation{
  \institution{University of Amsterdam}
  \city{Amsterdam}
  \country{The Netherlands}
}

\author{Jia-Hong Huang}
\orcid{0000-0001-7943-2591}
\affiliation{
  \institution{University of Amsterdam}
  \city{Amsterdam}
  \country{The Netherlands}
}

\author{Paul Groth}
\orcid{0000-0003-0183-6910}
\affiliation{
  \institution{University of Amsterdam}
  \city{Amsterdam}
  \country{The Netherlands}
}

\renewcommand{\shortauthors}{Anonymous Author(s)}

\begin{abstract}
Multi-Modal Knowledge Graphs (MMKGs) enrich entities with multiple modalities such as text and images, yet entities with highly similar multi-modal features remain difficult to distinguish. Temporal information of an entity can serve as an additional modality to disambiguate such entities, but existing approaches rarely treat time as a separate modality alongside text and images due to two major challenges: \textbf{(1) sparse temporal semantics}, which hinder alignment with richer modalities, and \textbf{(2) multiple timestamps}, which introduce noise or reduce robustness in representation learning. To address these challenges, we propose \textbf{Time Imprint}, a framework that treats time as an \emph{entity-level} modality and jointly aligns temporal, textual, and visual representations via a three-view contrastive objective. Additionally, to mitigate multi-timestamp ambiguity, Time Imprint studies a compact timestamp subset selection design space and aggregates the selected timestamps into a discriminative temporal embedding with attention pooling, balancing temporal specificity and robustness. Experiments on three MMKG benchmarks demonstrate that \textbf{Time Imprint} achieves state-of-the-art link prediction performance, improving Hits@1 by up to 6.07\% overall and yielding up to 58\% gains on the subset of the top-1\% ambiguity samples. We further examine different fusion strategies and the sensitivity to timestamp availability and quality, clarifying when and why time-as-modality is most beneficial, while adding only modest training overhead. We release our code at \url{https://anonymous.4open.science/r/Time-Imprint}.
\end{abstract}




\received{20 February 2007}
\received[revised]{12 March 2009}
\received[accepted]{5 June 2009}

\maketitle

\section{Introduction}

Multi-Modal Knowledge Graphs (MMKGs) enrich entity representations by integrating text, images, audio, and video with structured triples and are widely used in Web applications such as recommendation and search \cite{10.1145/3656579}. For example, in the case of a movie, an MMKG can augment triples with plot summaries, key frames, short clips, and audio capturing its visual style and pacing. 
These signals enable cross-modal alignment and yield more informative entity representations \cite{10.1145/3627673.3679545}, which is particularly valuable when constructing Knowledge Graphs (KGs) from noisy Web data.

\begin{figure}[t]
\centering
\includegraphics[width=\linewidth]{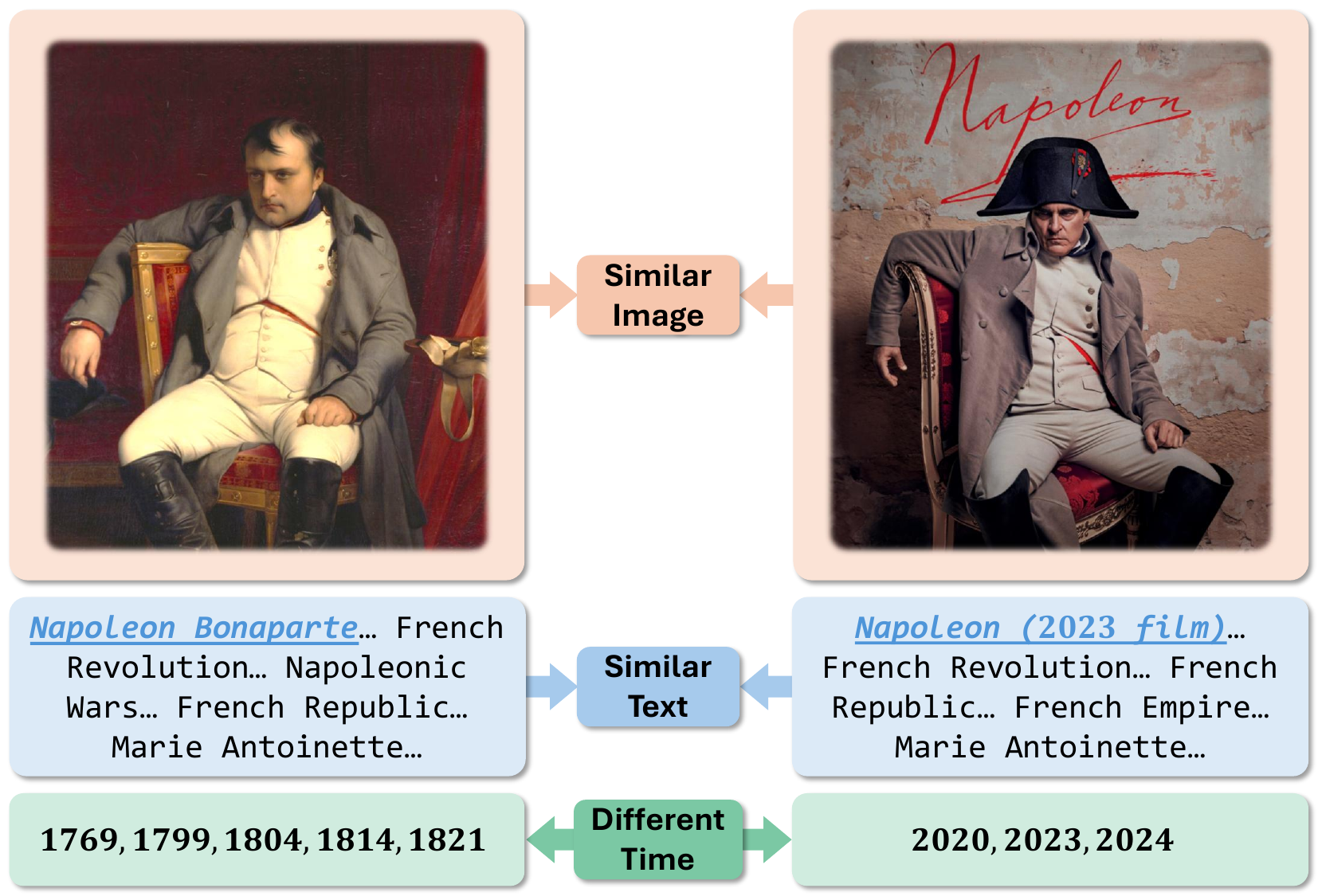}
\caption{``\texttt{Napoleon Bonaparte}'' (left) and ``\texttt{Napoleon (2023 film)}'' (right) share similar images and textual descriptions, which can mislead models. In contrast, their temporal information is different (1769-1821 vs. 2020-2024); therefore, incorporating time as a modality helps disambiguate them.}
\label{fig1}
\Description{fig1}
\end{figure}

These rich modalities enhance reasoning and disambiguation; however, challenges remain when different entities share highly similar text and visual content \cite{10.1145/3656579}. In such cases, temporal information can play a crucial role. Time is a fundamental attribute of knowledge and offers a powerful signal to distinguish between otherwise confusable entities \cite{cai2024surveytemporalknowledgegraph}. For example, Figure~\ref{fig1} compares ``\textbf{\texttt{Napoleon Bonaparte}}'' with ``\textbf{\texttt{Napoleon (2023 film)}}.'' Their images and textual descriptions overlap considerably, making them difficult to distinguish. However, temporal information about Bonaparte's lifetime (1769-1821) versus the film's production and release period (2020-2024) immediately highlights their differences. These cases suggest that temporal information should be modeled alongside text and images to resolve ambiguities among entities with highly similar multi-modal features.

Most existing approaches \cite{wang2023survey} treat time as an auxiliary label in temporal KG completion. In these methods, KG triples are extended into quadruples \textit{(head entity, relation, tail entity, timestamp)}, allowing link prediction to be conditioned on a specific temporal context \cite{song2024temporal}. Although this design facilitates temporal reasoning over relations, it restricts temporal information to the triple level and fails to enrich entity representations directly. In contrast, other modalities such as text and images are explicitly incorporated into entity-level representations in MMKGs \cite{wang2023survey}, yielding more informative and semantically grounded features. This mismatch, where time is modeled at the triple level while text and images are incorporated at the entity level, motivates a different perspective for MMKGs: modeling time as an additional \emph{entity-level} modality that can be aligned with text and images. Under this paradigm, however, two key challenges remain:

\textbf{Sparse Temporal Semantics:} Unlike images, text, audio, or video, each of which can independently identify an entity (e.g., a well-known singer can be recognized by their image, text description, audio, or video), time alone rarely provides sufficient information to characterize an entity \cite{9961954}. Consequently, temporal signals are inherently sparse and semantically weaker than other modalities. This property makes it difficult to align temporal features with richer modalities in a shared embedding space, making cross-modal alignment a primary challenge in temporal MMKGs \cite{cai2024surveytemporalknowledgegraph}.

\textbf{Multi-Timestamp Ambiguity:} Beyond alignment, modeling time as a separate modality introduces a second challenge: many entities are associated with multiple timestamps corresponding to different events (e.g., creation, publication, and award). Incorporating all timestamps can introduce noise and dilute distinctiveness, especially when minor or noisy events influence the aggregate representation. Conversely, selecting only a single timestamp increases specificity but reduces robustness to natural temporal variation. These different selection strategies, therefore produce systematically different representations. Managing the trade-off between temporal specificity and diversity thus becomes the second challenge.

To address these challenges, we propose \textbf{Time Imprint}, a framework that treats time as a distinct entity modality and aligns it with text and images through joint cross-modal training with a three-view contrastive objective. Specifically, to address \textbf{(1) Sparse Temporal Semantics}, we encode timestamps into temporal embeddings, inject them into multi-modal token representations, and learn aligned time-text-image features that make sparse temporal cues usable in the shared embedding space. To handle \textbf{(2) Multi-Timestamp Ambiguity}, we study a compact timestamp subset selection design space and aggregate the selected timestamps into a robust temporal embedding via attention pooling, explicitly characterizing the specificity-robustness trade-off.

We validate the design on three MMKG benchmarks. \textbf{Time Imprint} outperforms state-of-the-art models on the link prediction task, achieving gains of up to 6.07\% in Hits@1. On the subset of the top-1\% most ambiguous samples in each dataset, it improves Hits@1 by up to 58\%, highlighting the benefit of treating time as a separate modality. In addition, we analyze how performance varies under different fusion strategies and under missing or corrupted timestamps, clarifying the practical conditions under which time-as-modality provides the largest benefit. Our contributions are summarized as follows:
\begin{itemize}
    \item We propose \textbf{Time Imprint}, a framework that models time as an \emph{entity-level} modality and jointly aligns time, text, and image through a three-view contrastive objective, enabling systematic study of temporal injection and multi-view modality alignment in MMKGs.
    \item We represent entities with multiple timestamps by selecting a compact and informative subset and aggregating it with attention into a robust temporal embedding, thus balancing distinctiveness and robustness while analyzing how the number of timestamps influences model performance.
    \item Experiments on three public MMKG datasets show that our model consistently outperforms prior methods, improving link prediction Hits@1 by up to 6.07\%. On the subset of the top-1\% most ambiguous samples, it improves Hits@1 by up to 58\%, thereby demonstrating the effectiveness of modeling time as a separate modality.
\end{itemize}

\section{Related Work}

\subsection{Representation Learning in Multi-Modal Knowledge Graphs}

Multi-Modal Knowledge Graphs (MMKGs) enrich entity representations by integrating text and images with structured triples~\cite{ma2025transformerbasedmultimodalknowledgegraph,blp2021,mmkgprototypesWebconf}. Early models such as IKRL~\cite{10.5555/3172077.3172327} introduced visual embeddings, while subsequent work advanced modality fusion and alignment through contrastive learning (MCLEA~\cite{lin-etal-2022-multi}), optimal transport (OTKGE~\cite{NEURIPS2022_ffdb280e}), split-and-ensemble (MoSE~\cite{zhao-etal-2022-mose}), adaptive weighting (AdaMF~\cite{zhang-etal-2024-unleashing}), link-aware aggregation (LAFA~\cite{Shang_Zhao_Liu_Wang_2024}), visual entity alignment (EVA~\cite{Liu_Chen_Roth_Collier_2021}), hyperbolic alignment (HMEA~\cite{GUO2021598}), and noise-robust Transformer fusion (SNAG~\cite{chen-etal-2025-noise}). Among recent strong baselines, SNAG~\cite{chen-etal-2025-noise} employs a noise-tolerant Transformer that adaptively reweights modality tokens to suppress unreliable visual or textual signals, while MOMOK~\cite{zhang2025multiple} disentangles entity semantics into multiple relation-conditioned modality experts and fuses them through a mixture mechanism. Despite this progress, these approaches do not treat temporal information as a separate entity-level modality aligned with text and images, limiting their ability to disambiguate entities whose meaning depends on or evolves over time.

\subsection{Temporal Embedding in Knowledge Graphs}

Prior work typically encodes time as an auxiliary triple-level attribute. Existing approaches extend static models with temporal components~\cite{jiang-etal-2016-towards}, learn time-evolving representations (DE-SimplE~\cite{Goel_Kazemi_Brubaker_Poupart_2020}), apply sequence models (RE-Net~\cite{jin-etal-2020-recurrent}, CyGNet~\cite{Zhu_Chen_Fan_Cheng_Zhang_2021}), project onto time-dependent hyperplanes (HyTE~\cite{dasgupta-etal-2018-hyte}), combine temporal encoders with rule reasoning (TECHS~\cite{lin-etal-2023-techs}), adapt GNNs with recurrent units (REGCN~\cite{10.1145/3404835.3462963}), encode multi-granularity timestamps (LGRe~\cite{zhang-etal-2024-lgre}), and employ time-sensitive attention (TimeGate~\cite{11192270}). Some MMKG approaches incorporate time as literal features~\cite{wilcke2023endtoendlearningmultimodalknowledge}, but they lack cross-modal alignment or mechanisms to aggregate multiple timestamps. In contrast, our work models time as an independent modality, enabling explicit alignment with visual and textual features and robust aggregation of multiple timestamps per entity.

\subsection{Contrastive Learning in Multi-Modal Knowledge Graphs}

Contrastive objectives have been widely used to align structural, visual, and textual representations in MMKGs, including for entity alignment (MCLEA~\cite{lin-etal-2022-multi}), textual-visual alignment (CMR~\cite{10.1145/3626772.3657838}), CLIP-guided fusion (CLFA~\cite{zhang-etal-2024-multi}), and progressive modality consistency (PMF~\cite{huang-etal-2024-progressively}). However, existing methods focus on aligning dense modalities and do not treat time as a distinct modality. In contrast, our approach jointly aligns temporal, textual, and visual embeddings within a single contrastive framework.

\section{Methodology}
\label{sec:method}

\begin{figure*}[t]
\centering
\includegraphics[width=\textwidth]{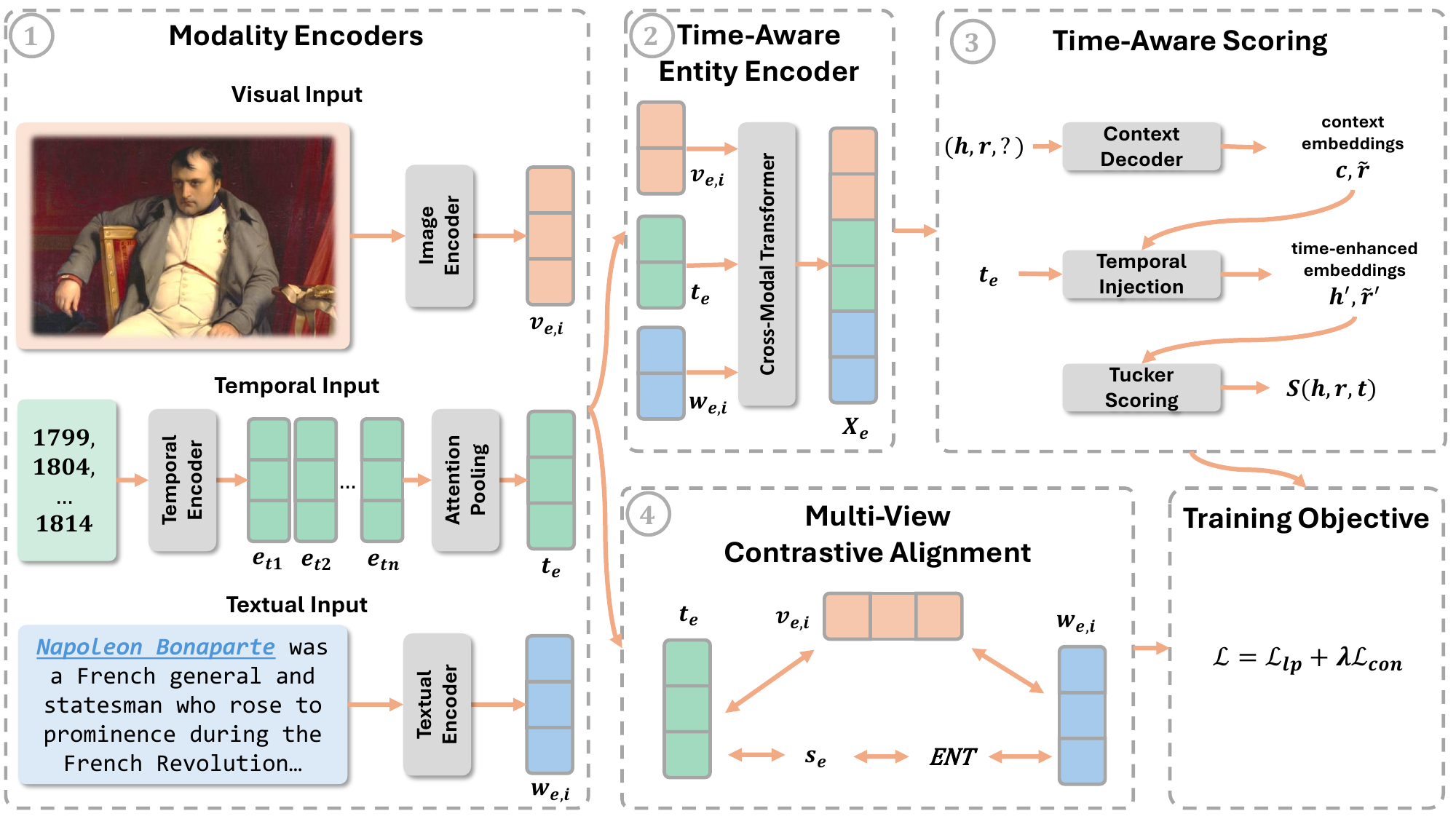}
\caption{Overview of Time Imprint. (1)~Modality encoders extract visual and textual tokens and aggregate selected timestamps into a temporal vector $\mathbf{t}_e$. (2)~The time-aware entity encoder prepends $\mathbf{t}_e$ as a prefix token and fuses all modalities via a cross-modal Transformer. (3)~Time-aware scoring injects $\mathbf{t}_e$ into the query decoder through a relation-aware gate and scores candidates using Tucker decomposition. (4)~Multi-view contrastive alignment aligns five entity views in a shared embedding space. The objectives $\mathcal{L}_{lp}$ and $\mathcal{L}_{con}$ are optimized jointly.}
\label{fig2}
\Description{fig2}
\end{figure*}

Figure~\ref{fig2} provides an overview of \textbf{Time Imprint}.
Our goal is to learn \emph{entity-level} time-aware representations in Multi-Modal Knowledge Graphs (MMKGs) by treating time as a distinct modality alongside text and images.

We denote an MMKG as $\mathcal{G}=(\mathcal{E},\mathcal{R},\mathcal{F})$, where $\mathcal{E}$ is the set of entities, $\mathcal{R}$ is the set of relations, and $\mathcal{F}\subseteq \mathcal{E}\times\mathcal{R}\times\mathcal{E}$ is the set of observed triples (facts).
Each entity $e\in\mathcal{E}$ is associated with a textual description $\mathcal{D}(e)$, a set of images $\mathcal{V}(e)$, and a set of candidate timestamps $\mathcal{Y}(e)=\{y_1,\dots,y_{m_e}\}$. 

\textcolor{black}{We adopt year-level granularity for two reasons. First, the temporal cues in our datasets are mainly years extracted from entity descriptions and image metadata (Section~\ref{sec:TimestampExtraction}); the available month- or day-level information mostly comes from media timestamps (e.g., the dates when photos were taken or uploaded) rather than from entity-level temporal facts. Second, finer granularity has little room for improvement: it can only affect entity pairs whose year ranges touch at a single boundary year (e.g., one entity active 1945--2006 and another active 2006--2026), where month-level information within the shared year may still separate them. Such boundary cases account for only $1.1\%$--$2.7\%$ of confusable entity pairs ($2.7\%$--$3.4\%$ over all pairs); among these cases with month-level data, month information separates none of the pairs in the ambiguity subsets. The potential benefit of finer granularity is thus bounded by less than $1\%$ of entity pairs, while year-level timestamps cover $86\%$--$95\%$ of entities. All remaining ambiguous pairs are either already separated at year level or overlap by multiple years, where no granularity can make their ranges disjoint; for these pairs, our model relies on differences in timestamp distributions rather than on finer time resolution.}

\subsection{Modality Encoders}
\label{sec:encoders}

\noindent \textbf{Temporal Encoder.}
We map each timestamp $y$ (a year) to a dense vector using a sinusoidal encoder~\cite{lin-etal-2023-techs}:  
{\small
\begin{equation}
\label{eq:time-encode}
\mathbf{e}_y \;=\; \sqrt{\tfrac{1}{d}}\;\big[\cos(\omega_1 y + \phi_1),\;\dots,\;\cos(\omega_{d} y + \phi_{d})\big],
\end{equation}}
where $\{\omega_i,\phi_i\}_{i=1}^{d}$ are learnable frequency and phase parameters and $d$ is the embedding dimension.
The sinusoidal form produces smoothly varying, temporal features: nearby years receive similar embeddings, while distant years are well separated.
Importantly, this encoder is applied once during initialization to produce entity-level temporal embeddings (described below), which are then stored as a fixed buffer. Downstream modules that consume these embeddings contain their own trainable parameters.

\noindent \textbf{Multi-Timestamp Selection and Attention Pooling.}
Entities often have multiple timestamps corresponding to different events (e.g., creation, publication, award).
Using all timestamps may introduce noise from minor or incidental events, while using only a single timestamp may be fragile.
We therefore select a compact subset $\mathcal{S}_K(e)\subseteq \mathcal{Y}(e)$ and aggregate it into a single temporal embedding via attention pooling.

\textit{Timestamp subset selection.}
We select at most $K$ timestamps that are closest to a center statistic of the entity's timestamp set.
Let $\mathrm{center}(\mathcal{Y}(e))$ denote a statistic such as the \emph{median}, \emph{earliest}, or \emph{latest} year. The selected subset is:
{\small
\begin{equation}
\label{eq:time-select}
\mathcal{S}_K(e)=\operatorname*{arg\,top\text{-}K}_{y\,\in\,\mathcal{Y}(e)}\!\Big(-\big|y-\mathrm{center}\big(\mathcal{Y}(e)\big)\big|\Big).
\end{equation}}
To preserve early historical evidence, we enforce an \emph{anchor} constraint: the earliest year in $\mathcal{Y}(e)$ is always included in $\mathcal{S}_K(e)$, replacing the least central element if necessary.
Our default configuration uses \emph{median-$K$} selection with the earliest-year anchor; we analyze alternative strategies in experiments.

\textit{Attention pooling.}
Each selected timestamp $y \in \mathcal{S}_K(e)$ is encoded using Eq.~\eqref{eq:time-encode}.
We aggregate the resulting vectors into a single temporal embedding using cosine-similarity-based attention:
{\small
\begin{equation}
\label{eq:time-attn}
\begin{aligned}
\boldsymbol{\mu} &= \frac{1}{|\mathcal{S}_K(e)|}\sum_{y}\mathbf{e}_y, \\
\alpha_y &= \frac{\exp\!\big(\mathrm{cos}(\mathbf{e}_y,\,\boldsymbol{\mu})\,/\,\tau_a\big)}{\sum_{y'}\exp\!\big(\mathrm{cos}(\mathbf{e}_{y'},\,\boldsymbol{\mu})\,/\,\tau_a\big)}, \\
\hat{\mathbf{e}}^{\,\mathrm{time}}_e &= \sum_{y\,\in\,\mathcal{S}_K(e)} \alpha_y\,\mathbf{e}_y.
\end{aligned}
\end{equation}
where $\mathrm{cos}(\cdot,\cdot)$ denotes cosine similarity, $\boldsymbol{\mu}$ is the mean embedding of the selected timestamps, and $\tau_a$ is a temperature that controls the sharpness of the attention distribution.
Timestamps whose embeddings lie close to the set centroid receive higher weights, naturally suppressing temporal outliers.

\noindent \textbf{Image and Text Encoders.}
Following prior MMKG work~\cite{DBLP:conf/aaai/ZhangCGXHLZC25}, we represent each modality as a set of discrete tokens.
For images, an image tokenizer (BEiT~\cite{peng2022beitv2maskedimage}) maps $\mathcal{V}(e)$ to visual tokens $\{\mathbf{v}_{e,1},\dots,\mathbf{v}_{e,M}\}$.
For text, a text tokenizer (BERT~\cite{devlin-etal-2019-bert}) maps $\mathcal{D}(e)$ to textual tokens $\{\mathbf{w}_{e,1},\dots,\mathbf{w}_{e,N}\}$.
Both token sets are truncated to the most frequent tokens per entity to control sequence length and reduce noise~\cite{DBLP:conf/aaai/ZhangCGXHLZC25}.
The corresponding token embeddings are obtained from pretrained embedding tables and kept frozen during training; only the downstream projection layers are trainable.

\subsection{Time-Aware Entity Encoder}
\label{sec:entity-encoder}

Given the modality-specific tokens and the aggregated temporal embedding $\hat{\mathbf{e}}^{\,\mathrm{time}}_e$, we construct a unified entity representation that integrates structural, visual, textual, and temporal information.

\noindent \textbf{Temporal prefix token.}
To inject temporal information into the entity encoder, we project the temporal embedding into a dedicated \emph{time token} and prepend it to the input sequence:
{\small
\begin{equation}
\label{eq:time-token}
\mathbf{t}_e = f_{\mathrm{time}}\!\big(\hat{\mathbf{e}}^{\,\mathrm{time}}_e\big),
\end{equation}}
where $f_{\mathrm{time}}$ is a linear projection initialized to zero, so that the time token has no effect at the start of training and its influence grows gradually as the model learns to exploit temporal cues.
This design avoids destabilizing early training while allowing the Transformer's self-attention to learn flexible interactions between the time token and other modality tokens.

\noindent \textbf{Cross-modal fusion.}
We form the input sequence by concatenating the entity token, time token, structural embedding, visual tokens, and textual tokens:
{\small
\begin{equation}
\label{eq:seq}
\mathcal{X}(e) = \big[\,\underbrace{[\mathrm{ENT}]}_{\text{cls}},\;\underbrace{\mathbf{t}_e}_{\text{time}},\;\underbrace{\mathbf{s}_e}_{\text{struct}},\;\underbrace{\mathbf{v}_{e,1},\dots,\mathbf{v}_{e,M}}_{\text{visual}},\;\underbrace{\mathbf{w}_{e,1},\dots,\mathbf{w}_{e,N}}_{\text{textual}}\,\big],
\end{equation}}
where $\mathbf{s}_e$ is a learnable structural embedding for entity $e$, and all tokens are projected to a shared dimension $d$.
Modality-specific positional embeddings and layer normalization are applied to visual and textual tokens before concatenation; entities lacking visual or textual tokens are handled via padding masks.
The sequence $\mathcal{X}(e)$ is processed by a multi-layer Transformer encoder with multi-head self-attention, in which every token (the time, structural, visual, and textual tokens) attends to all others, allowing the model to fuse the modalities and let the temporal token directly shape the cross-modal representation. The output at the $[\mathrm{ENT}]$ position serves as the unified entity representation:
{\small
\begin{equation}
\label{eq:entity-rep}
\mathbf{e} = \mathrm{Transformer}\!\big(\mathcal{X}(e)\big)_{[\mathrm{ENT}]}.
\end{equation}}

By placing the time token immediately after $[\mathrm{ENT}]$, the self-attention mechanism can directly attend to temporal information when forming the entity representation, enabling the model to condition entity semantics on time.

\subsection{Time-Aware Scoring}
\label{sec:scoring}

Given entity representations $\mathbf{e}$ from Eq.~\eqref{eq:entity-rep} and relation embeddings $\mathbf{r}$, obtained from a learnable relation embedding table indexed by relation id and randomly initialized then trained end-to-end with the rest of the model, we score candidate triples $(h, r, t)$ through a three-step process that further incorporates temporal information during the scoring stage.

\noindent \textbf{Context decoding.}
For a query such as $(h, r, ?)$, we construct a sequence $[\mathbf{h}, \mathbf{r}, \mathbf{?}]$ (with positional embeddings) and pass it through a Transformer decoder to produce contextualized representations.
The decoder output at the masked position yields the context embedding $\mathbf{c}$, and the output at the relation position yields the contextualized relation embedding $\tilde{\mathbf{r}}$:
{\small
\begin{equation}
\label{eq:decoder}
[\mathbf{c},\,\tilde{\mathbf{r}},\,\_\,] = \mathrm{Decoder}\!\big([\mathbf{h}+\mathbf{p}_h,\;\mathbf{r}+\mathbf{p}_r,\;\mathbf{?}+\mathbf{p}_t]\big),
\end{equation}}
where $\mathbf{p}_h, \mathbf{p}_r, \mathbf{p}_t$ are learnable positional embeddings for the head, relation, and tail positions, and the context embedding $\mathbf{c}$ is taken from the position corresponding to the masked entity (head or tail).

\noindent \textbf{Temporal injection.}
We inject the temporal embedding of the head entity into the scoring computation via a \emph{relation-aware gate} that modulates the injection strength based on the relation type:
{\small
\begin{equation}
\label{eq:time-inject}
\mathbf{g} = \sigma\!\big(W_g\,\tilde{\mathbf{r}}\big), \qquad
\mathbf{h}' = \mathbf{h} + \alpha\,\mathbf{g} \odot \hat{\mathbf{e}}^{\,\mathrm{time}}_h,
\end{equation}}
where $\sigma$ is the sigmoid function, $W_g$ is a learnable matrix (initialized to zero so that the gate starts at 0.5), $\alpha$ is a fixed scaling factor, and $\odot$ denotes element-wise multiplication.
The gate allows the model to learn which relations benefit from temporal information (e.g., \texttt{performer}, \texttt{cast\_member}) and which are time-agnostic (e.g., \texttt{has\_genre}).
The gated temporal embedding is added to the head entity embedding before scoring, enriching the head representation with time-dependent context.

\noindent \textbf{Temporal relation modulation.}
Beyond injecting time into entity embeddings, we also modulate the relation embedding using the temporal signal of the \emph{known} entity in the query.
For a tail query $(h, r, ?)$, the known entity is $h$; for a head query $(?, r, t)$, it is $t$.
We apply a learned projection:
{\small
\begin{equation}
\label{eq:rel-mod}
\tilde{\mathbf{r}}' = \tilde{\mathbf{r}} + \tanh\!\big(W_r\,\hat{\mathbf{e}}^{\,\mathrm{time}}_{\mathrm{known}}\big),
\end{equation}}
where $W_r$ is initialized to zero for a stable training start.
This modulation shifts the relation embedding toward a time-specific operating point, allowing the model to capture that the same relation may connect to different entity types in different time periods.

\noindent \textbf{Tucker scoring.}
We combine the context embedding and the modulated relation embedding using Tucker decomposition~\cite{balazevic-etal-2019-tucker} to score all candidate entities:
{\small
\begin{equation}
\label{eq:tucker}
\mathcal{S}(h,r,t) = \big(\mathcal{W} \times_1 \mathbf{c} \times_2 \tilde{\mathbf{r}}'\big)\,\mathbf{e}_t^\top,
\end{equation}}
where $\mathcal{W}$ is a learnable core tensor.
Scoring is performed over all candidate entities simultaneously for efficient training and evaluation.

\noindent \textbf{Link prediction loss.}
We adopt the standard softmax cross-entropy loss over corrupted heads and tails:
{\small
\begin{equation}
\label{eq:lp-loss}
\mathcal{L}_{\mathrm{lp}} = -\!\!\sum_{(h,r,t)\in\mathcal{F}}\!\bigg[\log\frac{\exp\big(\mathcal{S}(h,r,t)\big)}{\sum_{t'}\exp\big(\mathcal{S}(h,r,t')\big)} + \log\frac{\exp\big(\mathcal{S}(h,r,t)\big)}{\sum_{h'}\exp\big(\mathcal{S}(h',r,t)\big)}\bigg].
\end{equation}}

\subsection{Multi-View Contrastive Alignment}
\label{sec:contrastive}

To explicitly align temporal, visual, and textual representations in a shared embedding space, we introduce a multi-view contrastive objective.
For each entity $e$, we construct five views from the entity encoder output (Eq.~\eqref{eq:entity-rep}):
(i)~the $[\mathrm{ENT}]$ representation $\mathbf{e}$,
(ii)~a \emph{global} view from mean-pooling the full token sequence,
(iii)~a \emph{visual} view from mean-pooling the visual token outputs,
(iv)~a \emph{textual} view from mean-pooling the textual token outputs,
and (v)~a \emph{temporal} view $\hat{\mathbf{e}}^{\,\mathrm{time}}_e$ from the aggregated temporal embedding.
We denote these views as $\mathcal{C}(e) = \{\mathbf{e},\,\mathbf{e}^{\mathrm{glob}},\,\mathbf{e}^{\mathrm{vis}},\,\mathbf{e}^{\mathrm{txt}},\,\hat{\mathbf{e}}^{\,\mathrm{time}}_e\}$.

We use an InfoNCE-style loss  that encourages views of the same entity to be close and views from different entities to be far apart:
{\small
\begin{equation}
\label{eq:con}
\mathcal{L}_{\mathrm{con}} = \sum_{e\in\mathcal{B}}\;\sum_{\mathbf{a}\in\mathcal{C}(e)}\;\sum_{\mathbf{p}\in\mathcal{C}(e)\setminus\{\mathbf{a}\}} -\log\frac{\exp\!\big(\mathrm{cos}(\mathbf{a},\mathbf{p})/\tau\big)}{\sum_{e'\in\mathcal{B}}\sum_{\mathbf{n}\in\mathcal{C}(e')}\exp\!\big(\mathrm{cos}(\mathbf{a},\mathbf{n})/\tau\big)},
\end{equation}}
where $\mathcal{B}$ is a randomly sampled subset of entities from the minibatch, $\mathrm{cos}(\cdot,\cdot)$ denotes cosine similarity, and $\tau$ is a temperature parameter.
The temporal view participates in all pairwise combinations within $\mathcal{C}(e)$, with its contribution further weighted by a factor $\gamma$ to control the relative emphasis on temporal alignment.
This objective encourages the sparse temporal embedding to become comparable with richer visual and textual features in the shared embedding space, directly addressing the \textit{sparse temporal semantics} challenge.

\subsection{Training Objective}
\label{sec:objective}

The final objective jointly optimizes link prediction and multi-view alignment:
{\small
\begin{equation}
\label{eq:final}
\mathcal{L} = \mathcal{L}_{\mathrm{lp}} + \lambda\,\mathcal{L}_{\mathrm{con}},
\end{equation}}
where $\lambda$ controls the weight of contrastive loss.
By combining a supervised link prediction signal with a self-supervised temporal-visual-textual alignment signal, Time Imprint learns entity representations that are both structurally grounded and temporally discriminative.
The three temporal injection points-entity encoding (\S\ref{sec:entity-encoder}), scoring (\S\ref{sec:scoring}), and contrastive alignment (\S\ref{sec:contrastive})-operate at different levels of the model and provide complementary benefits, as we verify through ablation studies in \S\ref{sec:rq2}.

\section{Experiments}
\label{sec:exp}

This section addresses four research questions regarding modeling time as an entity-level modality in MMKGs.

\noindent \textbf{RQ1 (Overall Performance):} How does Time Imprint compare with state-of-the-art methods?

\noindent \textbf{RQ2 (Ablation Study):} What is the contribution of each temporal injection stage (entity encoding, scoring injection, and contrastive alignment)?

\noindent \textbf{RQ3 (Multi-Timestamp Modeling):} How do the number of selected timestamps and the selection strategy affect performance?

\noindent \textbf{RQ4 (Robustness):} How sensitive is Time Imprint to timestamp noise?

\subsection{Experimental Setup}
\label{sec:exp_setup}

\noindent \textbf{Datasets.}
We evaluate our approach on three widely used MMKG benchmarks: DB15K, MKG-W, and MKG-Y.
Each dataset provides entities with structural triples, textual descriptions, and associated images.
Across the three datasets, we obtain timestamps for 38,961 out of 42,842 entities (90.94\%), with an average of 3.72 candidate years per timestamped entity.
A manual audit of 100 randomly sampled entities confirms that 89\% contain at least one correct year, motivating our multi-timestamp selection and attention pooling strategy (\S\ref{sec:encoders}) to suppress noisy or incidental years.
Detailed dataset statistics are provided in Appendix Datasets.

\noindent \textbf{Evaluation protocol and metrics.}
We follow the standard filtered link prediction protocol and report Mean Reciprocal Rank (MRR) and Hits@$n$ ($n\!=\!1,3,10$), where higher values indicate better performance.
We highlight the best results in \textbf{bold} and the second-best with \underline{underline}.
Relative improvements:
{\small
\begin{equation}
\label{eq:improvement}
\text{Improvement (\%)} = \left( \frac{\text{Our Result} - \text{Previous Best Result}}{\text{Previous Best Result}} \right) \times 100\%.
\end{equation}}

\subsection{Timestamp Extraction}
\label{sec:TimestampExtraction}

For all three datasets, we extract temporal information for each entity using a unified two-step strategy. The first source of time information comes from the textual description of each entity. These descriptions, originally collected from DBpedia, often contain references to specific years. We apply a rule-based method to detect and extract all years mentioned in the text. For example, the entity ``\textbf{\texttt{Napoleon (2023 film)}}'' is described as ``\texttt{Napoleon is a 2023 epic historical war film co-produced and directed by Ridley Scott from a screenplay by David Scarpa}'', from which we extract ``2023'' as a valid year. If multiple years are present, we retain all of them in ascending order.

The second source is based on entity images. Using the existing entity identifiers provided in the datasets, we retrieve corresponding image resources and collect their available metadata, particularly the dates associated with when the photos were taken. These dates are extracted and stored as additional temporal signals. When multiple timestamps are available, we sort and preserve them in ascending order.

\begin{table*}[t]
\caption{Link prediction results on three MMKG benchmarks. We report MRR and Hits@$n$ ($n=1,3,10$). The best results are shown in \textbf{bold} and the second-best are \underline{underlined}.}
\label{table1}
\centering
\begin{tabular}{@{}c|cccc|cccc|cccc@{}}
\toprule
\multirow{2}{*}{\textbf{Model}}        & \multicolumn{4}{c|}{\textbf{DB15K} \cite{liu2019mmkg}}                               & \multicolumn{4}{c|}{\textbf{MKG-W} \cite{10.1145/3503161.3548388}}                               & \multicolumn{4}{c}{\textbf{MKG-Y} \cite{10.1145/3503161.3548388}}                                \\
                                       & \textbf{MRR}   & \textbf{H@1}   & \textbf{H@3}   & \textbf{H@10}  & \textbf{MRR}   & \textbf{H@1}   & \textbf{H@3}   & \textbf{H@10}  & \textbf{MRR}   & \textbf{H@1}   & \textbf{H@3}   & \textbf{H@10}  \\ \midrule
\textbf{TransE} \cite{NIPS2013_1cecc7a7}                        & 24.86          & 12.78          & 31.48          & 47.07          & 29.19          & 21.06          & 33.20          & 44.23          & 30.73          & 23.45          & 35.18          & 43.37          \\
\textbf{DistMult} \cite{yang2015embeddingentitiesrelationslearning}                      & 23.03          & 14.78          & 26.28          & 39.59          & 20.99          & 15.93          & 22.28          & 30.86          & 25.04          & 19.33          & 27.80          & 35.95          \\
\textbf{ComplEx} \cite{pmlr-v48-trouillon16}                       & 27.48          & 18.37          & 31.57          & 45.37          & 24.93          & 19.09          & 26.69          & 36.73          & 28.71          & 22.26          & 32.12          & 40.93          \\
\textbf{RotatE} \cite{sun2019rotateknowledgegraphembedding}                        & 29.28          & 17.87          & 36.12          & 49.66          & 33.67          & 26.80          & 36.68          & 46.73          & 34.95          & 29.10          & 38.35          & 45.30          \\
\textbf{TuckER} \cite{balazevic-etal-2019-tucker}                       & 33.86          & 25.33          & 37.91          & 50.38          & 30.39          & 24.44          & 32.91          & 41.25          & 37.05          & 34.59          & 38.43          & 41.45          \\
\textbf{IKRL} \cite{10.5555/3172077.3172327}                         & 26.82          & 14.09          & 34.93          & 49.09          & 32.36          & 26.11          & 34.75          & 44.07          & 33.22          & 30.37          & 34.28          & 38.26          \\
\textbf{TBKGC} \cite{mousselly2018multimodal}                        & 28.40          & 15.61          & 37.03          & 49.86          & 31.48          & 25.31          & 33.98          & 43.24          & 33.99          & 30.47          & 35.27          & 40.07          \\
\textbf{TransAE} \cite{8852079}                      & 28.09          & 21.25          & 31.17          & 41.17          & 30.00          & 21.23          & 34.91          & 44.72          & 28.10          & 25.31          & 29.10          & 33.03          \\
\textbf{MMKRL} \cite{lu2022mmkrl}                        & 26.81          & 13.85          & 35.07          & 49.39          & 30.10          & 22.16          & 34.09          & 44.69          & 36.81          & 31.66          & 39.79          & 45.31          \\
\textbf{RSME} \cite{wang2021visual}                         & 29.76          & 24.15          & 32.12          & 40.29          & 29.23          & 23.36          & 31.97          & 40.43          & 34.44          & 31.78          & 36.07          & 39.09          \\
\textbf{VBKGC} \cite{zhang2022knowledge}                        & 30.61          & 19.75          & 37.18          & 49.44          & 30.61          & 24.91          & 33.01          & 40.88          & 37.04          & 33.76          & 38.75          & 42.30          \\
\textbf{OTKGE} \cite{NEURIPS2022_ffdb280e}                        & 23.86          & 18.45          & 25.89          & 34.23          & 34.36          & 28.85          & 36.25          & 44.88          & 35.51          & 31.97          & 37.18          & 41.38          \\
\textbf{MACO} \cite{10.1007/978-3-031-44693-1_10}                         & 27.41          & 14.61          & 35.59          & 50.00          & 31.74          & 25.23          & 34.23          & 44.37          & 34.98          & 31.59          & 36.68          & 40.51          \\
\textbf{IMF} \cite{10.1145/3543507.3583554}                          & 32.25          & 24.20          & 36.00          & 48.19          & 34.50          & 28.77          & 36.62          & 45.44          & 35.79          & 32.95          & 37.14          & 40.63          \\
\textbf{QEB} \cite{wang2023tiva}                          & 28.18          & 14.82          & 36.67          & 51.55          & 32.38          & 25.47          & 35.06          & 45.32          & 34.37          & 29.49          & 36.95          & 42.32          \\
\textbf{VISTA} \cite{lee-etal-2023-vista}                        & 30.42          & 22.49          & 33.56          & 45.94          & 32.91          & 26.12          & 35.38          & 45.61          & 30.45          & 24.87          & 32.39          & 41.53          \\
\textbf{AdaMF} \cite{zhang-etal-2024-unleashing}                        & 32.51          & 21.31          & 39.67          & 51.68          & 34.27          & 27.21          & 37.86          & 47.21          & 38.06          & 33.49          & 40.44          & 45.48          \\
\textbf{MANS} \cite{10191314}                         & 28.82          & 16.87          & 36.58          & 49.26          & 30.88          & 24.89          & 33.63          & 41.78          & 29.03          & 25.25          & 31.35          & 34.49          \\
\textbf{MMRNS} \cite{10.1145/3503161.3548388}                        & 32.68          & 23.01          & 37.86          & 51.01          & 35.03          & 28.59          & 37.49          & 47.47          & 35.93          & 30.53          & 39.07          & 45.47          \\
\textbf{NativE} \cite{10.1145/3626772.3657800}                       & 37.16          & 28.01          & 41.31    & 54.13 & 36.58    & 29.56          & 38.97    & 48.27    & 39.04    & 34.79          & 40.93 & 45.88    \\
\textbf{MyGO} \cite{DBLP:conf/aaai/ZhangCGXHLZC25}                         & 37.72    & 30.08    & 41.26          & 52.21          & 36.10          & 29.78    & 38.54          & 47.75          & 38.44          & 35.01    & 39.84          & 44.19          \\
\textbf{M-REFT} \cite{10.1145/3726302.3730082}        & 37.60                & 30.30                & 41.00                & 51.50               & 37.20                   & {\ul 30.80}             & 39.90                   & 48.90                  & 37.30                   & 34.50                   & 39.20                  & 42.90                  \\
\textbf{SNAG} \cite{chen-etal-2025-noise}          & 36.30       & 27.40       & 41.10      & 53.00      & {\ul 37.30}             & 30.20                   & {\ul 40.50}             & {\ul 50.30}            & {\ul 39.50}             & {\ul 35.40}             & {\ul 41.10}            & \textbf{47.10}         \\
\textbf{MOMOK} \cite{zhang2025multiple}               & {\ul 39.57}          & \textbf{32.38}       & \textbf{43.45}       & \textbf{54.14}      & 35.89                   & 30.38                   & 38.67                   & 47.62                  & 37.91                   & 35.09                   & 40.21                  & 45.73                  \\ \midrule
\textbf{Time Imprint}                                 & \textbf{39.75}       & {\ul 31.76}          & {\ul 42.96}          & {\ul 53.86}         & \textbf{38.99}          & \textbf{32.67}          & \textbf{41.38}          & \textbf{50.46}         & \textbf{39.62}          & \textbf{35.92}          & \textbf{41.62}         & {\ul 46.58}            \\
\textit{improv. (\%)}                                 & +0.45                & -1.91                & -1.13                & -0.52               & +4.53                   & +6.07                   & +2.17                   & +0.32                  & +0.30                   & +1.47                   & +1.27                  & -1.10                  \\ \bottomrule
\end{tabular}
\end{table*}

\begin{table*}[t]
\caption{Link prediction results on the top-1\% multi-modal ambiguity subset ($\sim$150 entities per dataset with highest text-image cosine similarity). Time Imprint outperforms the state-of-the-art MMKG models SNAG and MOMOK, with improvements of up to +58.21\% in Hits@1 on MKG-W.}
\label{table_similar}
\centering
\begin{tabular}{@{}c|cccc|cccc|cccc@{}}
\toprule
\multirow{2}{*}{\textbf{Model}} & \multicolumn{4}{c|}{\textbf{DB15K}}                               & \multicolumn{4}{c|}{\textbf{MKG-W}}                               & \multicolumn{4}{c}{\textbf{MKG-Y}}                                \\
                                & \textbf{MRR}   & \textbf{H@1}   & \textbf{H@3}   & \textbf{H@10}  & \textbf{MRR}   & \textbf{H@1}   & \textbf{H@3}   & \textbf{H@10}  & \textbf{MRR}   & \textbf{H@1}   & \textbf{H@3}   & \textbf{H@10}  \\ \midrule
\textbf{SNAG}  \cite{chen-etal-2025-noise}                  & 29.26          & 21.33          & 37.33          & 52.56          & 29.60          & 22.30          & 28.33          & 47.60          & 40.65          & 33.96          & 42.65          & 45.70          \\
\textbf{MOMOK}  \cite{zhang2025multiple}                 & 28.37          & 22.29          & 36.53          & 53.83          & 29.12          & 21.33          & 29.72          & 48.22          & 36.81          & 31.72          & 41.07          & 47.57          \\ \midrule
\textbf{Time Imprint}           & \textbf{36.36} & \textbf{28.53} & \textbf{45.17} & \textbf{56.46} & \textbf{41.03} & \textbf{35.28} & \textbf{44.52} & \textbf{53.73} & \textbf{43.72} & \textbf{38.36} & \textbf{44.83} & \textbf{49.37} \\
\textit{improv. (\%)}           & +24.27         & +27.99         & +21.00         & +4.89          & +38.61         & +58.21         & +49.80         & +11.43         & +7.55          & +12.96         & +5.11          & +3.78          \\ \bottomrule
\end{tabular}
\end{table*}

\subsection{Baselines}
\label{sec:Baselines}

We compare against representative KG embedding and MMKG completion methods, including TransE~\cite{NIPS2013_1cecc7a7}, DistMult~\cite{yang2015embeddingentitiesrelationslearning}, ComplEx~\cite{pmlr-v48-trouillon16}, RotatE~\cite{sun2019rotateknowledgegraphembedding}, TuckER~\cite{balazevic-etal-2019-tucker}. For MMKG completion, we consider IKRL~\cite{10.5555/3172077.3172327}, TBKGC~\cite{mousselly2018multimodal}, TransAE~\cite{8852079}, MMKRL~\cite{lu2022mmkrl}, RSME~\cite{wang2021visual}, VBKGC~\cite{zhang2022knowledge}, OTKGE~\cite{NEURIPS2022_ffdb280e}, MACO~\cite{10.1007/978-3-031-44693-1_10}, IMF~\cite{10.1145/3543507.3583554}, QEB~\cite{wang2023tiva}, VISTA~\cite{lee-etal-2023-vista}, AdaMF~\cite{zhang-etal-2024-unleashing}, NativE~\cite{10.1145/3626772.3657800}, MyGO~\cite{DBLP:conf/aaai/ZhangCGXHLZC25}, M-REFT~\cite{10.1145/3726302.3730082}, SNAG~\cite{chen-etal-2025-noise}, MOMOK~\cite{zhang2025multiple}.
Our implementation is built on the MyGO framework.\footnote{\url{https://github.com/zjukg/MyGO}}

\subsection{Training Details}

We build on the MyGO framework \cite{DBLP:conf/aaai/ZhangCGXHLZC25} and adopt the BERT and BEIT tokenizers for text and images, with vocabulary sizes of 32,000 and 8,192, respectively. Models are trained for 1,500 epochs using the Adam optimizer, with a batch size of 1,024 and an embedding dimension of 256. All experiments are conducted on a single NVIDIA H100 GPU.

\subsection{Overall Performance (RQ1)}
\label{sec:rq1}

\noindent \textbf{Overall performance.}
Table~\ref{table1} reports link prediction results on DB15K, MKG-W, and MKG-Y.
Time Imprint achieves the best overall performance across the three datasets.
On DB15K, it obtains the highest MRR (39.75, +0.45\% over the previous best MOMOK) while remaining competitive on other metrics.
On MKG-W, Time Imprint establishes a new state-of-the-art across all four metrics: MRR reaches 38.99 (+4.53\%), Hits@1 increases to 32.67 (+6.07\%), Hits@3 to 41.38 (+2.17\%), and Hits@10 to 50.46 (+0.32\%).
On MKG-Y, Time Imprint achieves the best MRR (39.62, +0.30\%), Hits@1 (35.92, +1.47\%), and Hits@3 (41.62, +1.27\%), while Hits@10 (46.58) ranks second to SNAG (47.10).
The largest improvements appear on MKG-W, where Time Imprint improves Hits@1 by over 6\%, suggesting that temporal information is most beneficial when it complements existing modalities to resolve entity ambiguity.

\noindent \textbf{Why time helps.}
We attribute these gains to the three-stage temporal injection design.
First, the temporal prefix token in the entity encoder allows the Transformer to attend directly to temporal cues when forming entity representations.
Second, the gated temporal injection and relation modulation in the scoring stage enable time-conditioned predictions, particularly for temporally sensitive relations.
Third, multi-view contrastive alignment encourages the sparse temporal embedding to become comparable with other modalities in the shared embedding space.

\noindent \textbf{Multi-modal ambiguity subset.}
Overall averages can mask the most challenging disambiguation cases where temporal information is most informative.
Following the intuition in Figure~\ref{fig1}, we construct a challenging subset for each dataset by selecting the top 1\% most ambiguous entities, defined as those with the highest average nearest-neighbor cosine similarity in the pretrained text and image embedding spaces.
This yields approximately 150 entities per dataset whose multi-modal features are highly confusable, and we evaluate all test triples involving these entities.

We restrict the comparison on this subset to SNAG and MOMOK for two reasons. First, they are the two most recent multi-modal baselines and are the strongest overall in Table~\ref{table1}: MOMOK attains the best results on DB15K (e.g., the highest Hits@1 and Hits@3), while SNAG is the strongest on MKG-W and MKG-Y (e.g., the best Hits@10 on both), so together they represent the state-of-the-art across all three datasets. Second, and more importantly for this experiment, both are explicitly designed to handle the difficulty that defines this subset: SNAG uses noise-tolerant modality fusion and MOMOK uses relation-conditioned modality experts, both targeting entities whose text and image signals are hard to separate. They are therefore the most relevant and most competitive references for measuring how well a model disambiguates the top-1\% most ambiguous entities, where visual and textual cues alone are least reliable. Table~\ref{table_similar} reports this comparison.

Time Imprint substantially outperforms both across all three datasets.
The gains are most pronounced on MKG-W, where MRR increases by +38.61\% over the best baseline, Hits@1 by +58.21\%, and Hits@3 by +49.80\%.
On DB15K, Time Imprint improves MRR by +24.27\% and Hits@1 by +27.99\%.
On MKG-Y, where the baselines already perform strongly on this subset, Time Imprint still achieves consistent gains (MRR +7.55\%, Hits@1 +12.96\%).
These results confirm that temporal information is most valuable precisely when visual and textual features alone cannot distinguish entities, validating the core motivation of treating time as a distinct modality.

\begin{table}[t]
\caption{Ablation study of the three temporal injection stages. Each variant removes one component while keeping the other two active. Full Time Imprint~(v) achieves the best results on all datasets. $\Delta$ values are relative to~(v).}
\label{table_ablation}
\centering
\scriptsize
\begin{tabular}{@{}cl|cccc@{}}
\toprule
& \textbf{Variant} & \textbf{MRR} & \textbf{H@1} & \textbf{H@3} & \textbf{H@10} \\ \midrule
\multirow{5}{*}{\rotatebox{90}{\textbf{DB15K}}}
& (i) Baseline            & 38.33 \tiny{($-$1.42)} & 30.66 \tiny{($-$1.10)} & 42.15 \tiny{($-$0.81)} & 53.13 \tiny{($-$0.73)} \\
& (ii) Disable Token       & 38.37 \tiny{($-$1.38)} & 30.76 \tiny{($-$1.00)} & 42.00 \tiny{($-$0.96)} & 53.18 \tiny{($-$0.68)} \\
& (iii) Disable Scoring     & 38.43 \tiny{($-$1.32)} & 30.80 \tiny{($-$0.96)} & 41.94 \tiny{($-$1.02)} & 53.38 \tiny{($-$0.48)} \\
& (iv) Disable Contrastive & 38.55 \tiny{($-$1.20)} & 30.94 \tiny{($-$0.82)} & 42.02 \tiny{($-$0.94)} & 53.16 \tiny{($-$0.70)} \\
& \textbf{(v) Full}       & \textbf{39.75} & \textbf{31.76} & \textbf{42.96} & \textbf{53.86} \\ \midrule
\multirow{5}{*}{\rotatebox{90}{\textbf{MKG-W}}}
& (i) Baseline            & 38.02 \tiny{($-$0.97)} & 31.55 \tiny{($-$1.12)} & 40.99 \tiny{($-$0.39)} & 49.72 \tiny{($-$0.74)} \\
& (ii) Disable Token       & 38.24 \tiny{($-$0.75)} & 31.96 \tiny{($-$0.71)} & 40.26 \tiny{($-$1.12)} & 50.04 \tiny{($-$0.42)} \\
& (iii) Disable Scoring     & 38.04 \tiny{($-$0.95)} & 31.60 \tiny{($-$1.07)} & 40.37 \tiny{($-$1.01)} & 50.34 \tiny{($-$0.12)} \\
& (iv) Disable Contrastive & 38.57 \tiny{($-$0.42)} & 32.22 \tiny{($-$0.45)} & 41.00 \tiny{($-$0.38)} & 50.36 \tiny{($-$0.10)} \\
& \textbf{(v) Full}       & \textbf{38.99} & \textbf{32.67} & \textbf{41.38} & \textbf{50.46} \\ \midrule
\multirow{5}{*}{\rotatebox{90}{\textbf{MKG-Y}}}
& (i) Baseline            & 38.72 \tiny{($-$0.90)} & 35.47 \tiny{($-$0.45)} & 39.88 \tiny{($-$1.74)} & 44.42 \tiny{($-$2.16)} \\
& (ii) Disable Token       & 39.06 \tiny{($-$0.56)} & 35.47 \tiny{($-$0.45)} & 40.52 \tiny{($-$1.10)} & 45.64 \tiny{($-$0.94)} \\
& (iii) Disable Scoring     & 38.67 \tiny{($-$0.95)} & 35.32 \tiny{($-$0.60)} & 39.86 \tiny{($-$1.76)} & 44.39 \tiny{($-$2.19)} \\
& (iv) Disable Contrastive & 38.99 \tiny{($-$0.63)} & 35.43 \tiny{($-$0.49)} & 40.50 \tiny{($-$1.12)} & 45.32 \tiny{($-$1.26)} \\
& \textbf{(v) Full}       & \textbf{39.62} & \textbf{35.92} & \textbf{41.62} & \textbf{46.58} \\ \bottomrule
\end{tabular}
\end{table}

\subsection{Ablation Study (RQ2)}
\label{sec:rq2}

Time Imprint injects temporal information at three stages: (1) the temporal prefix token in the entity encoder (\S\ref{sec:entity-encoder}), (2) the gated injection and relation modulation in the scoring decoder (\S\ref{sec:scoring}), and (3) the temporal view in contrastive alignment (\S\ref{sec:contrastive}).
To quantify the contribution of each stage, we train five variants on all three datasets, removing one component at a time while keeping all other hyper-parameters fixed:

(i) Baseline: all three temporal components were removed.

(ii) Disable Time Token: scoring injection and contrastive time view remain active, but the temporal prefix token is replaced with a zero vector.

(iii) Disable Scoring: entity encoding and contrastive time view remain active, but temporal injection in the scoring decoder is disabled.

(iv) Disable Contrastive: entity encoding and scoring injection remain active, but the temporal view weight in the contrastive loss is set to zero.

(v) Full Time Imprint: all three temporal components active.

Table~\ref{table_ablation} reports the results, from which three findings emerge.
First, the no-time Baseline~(i) is among the weakest configurations and is the weakest on MRR and Hits@1 for DB15K and MKG-W, confirming the overall benefit of temporal information.
On MKG-W, Full Time Imprint~(v) improves over Baseline~(i) by +1.12 in Hits@1 and +0.97 in MRR; on DB15K the gains are even larger reaching +1.10 Hits@1 and +1.42 MRR.
Second, scoring injection is the most critical single component: removing it (variant~iii) causes the largest Hits@1 drop on MKG-W ($-$1.07) and the largest MRR drop on MKG-Y ($-$0.95), indicating that time-conditioned scoring has the most direct impact on ranking quality.

Third, the three components are interdependent rather than purely additive. On MRR and Hits@1, removing all three (Baseline~i) is usually the most damaging configuration, especially on DB15K and MKG-W, which indicates that the stages largely reinforce one another. However, this is not uniform: on Hits@3 across all datasets, and for most metrics on MKG-Y, removing a single component, most often the scoring injection (iii), degrades performance \emph{below} the no-time Baseline~(i). We attribute this to interaction effects between the stages: keeping only a subset of the temporal components forces the model to rely on a partially-injected temporal signal that is inconsistent across encoding, scoring, and contrastive alignment, which can be more harmful than removing temporal information altogether. This interdependence explains why the full model (v), where all three stages are mutually consistent, is the only configuration that is best on every dataset and metric.

The contrastive temporal view (iv) provides the smallest individual gain but remains consistently beneficial across datasets, acting as a complementary self-supervised alignment signal.

\subsection{Multi-Timestamp Modeling (RQ3)}
\label{sec:rq3}

\begin{figure}[t]
\centering
\includegraphics[width=\linewidth]{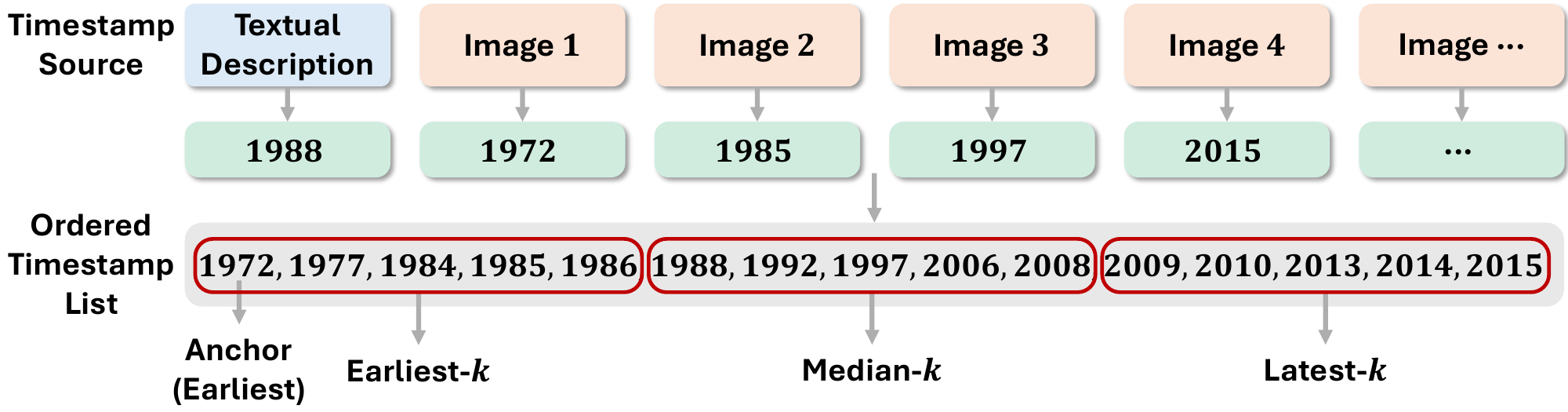}
\caption{Illustration of timestamp extraction and selection strategies. For each entity, timestamps are collected from textual descriptions and image metadata, and then sorted in ascending order from the earliest to the most recent year.}
\label{fig4}
\Description{fig4}
\end{figure}

\begin{figure*}[htb]
\centering
\subfigure[DB15K Hits@1]{\includegraphics[width=0.32\linewidth]{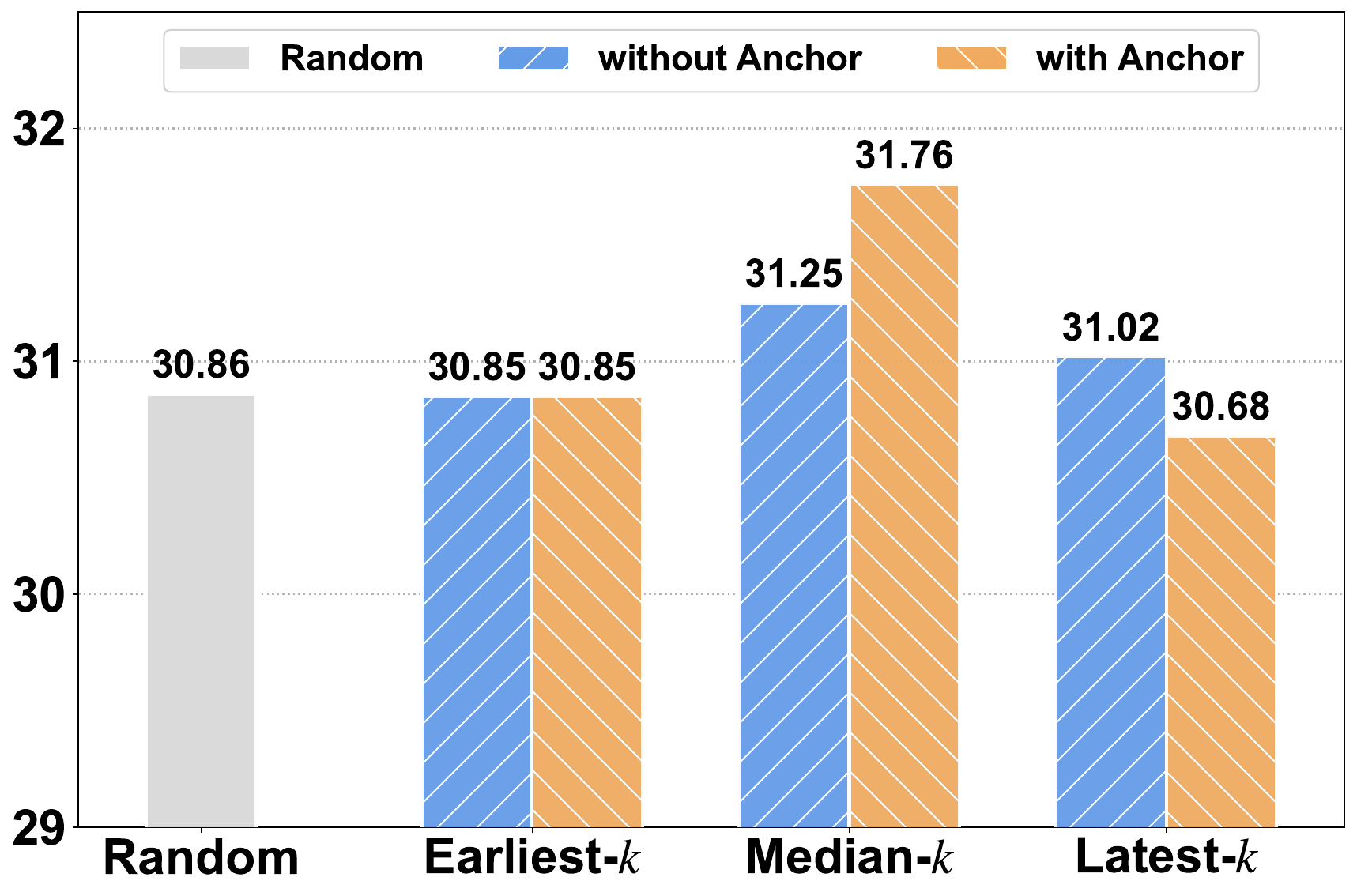}}
\subfigure[MKG-W Hits@1]{\includegraphics[width=0.32\linewidth]{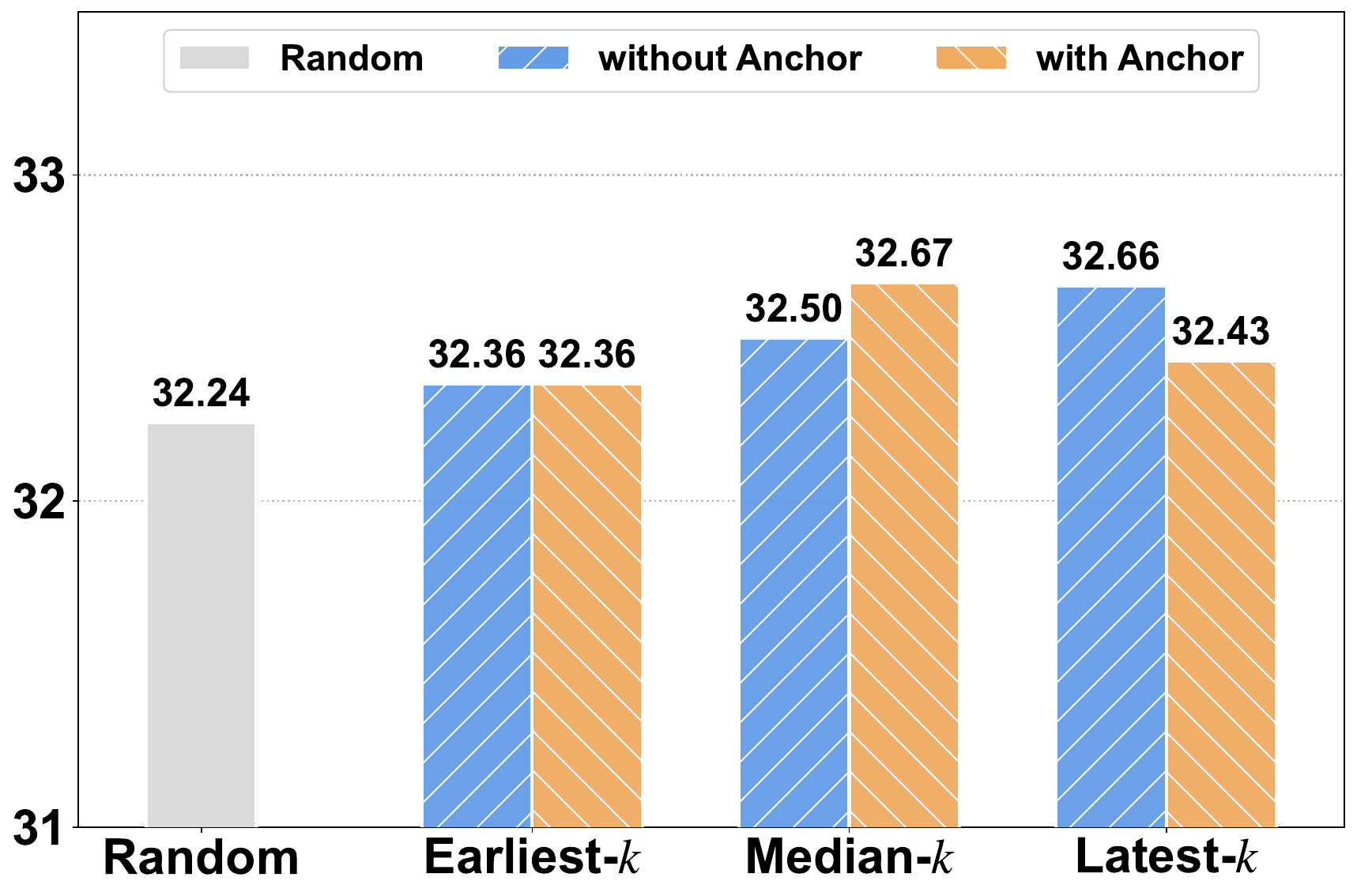}}
\subfigure[MKG-Y Hits@1]{\includegraphics[width=0.32\linewidth]{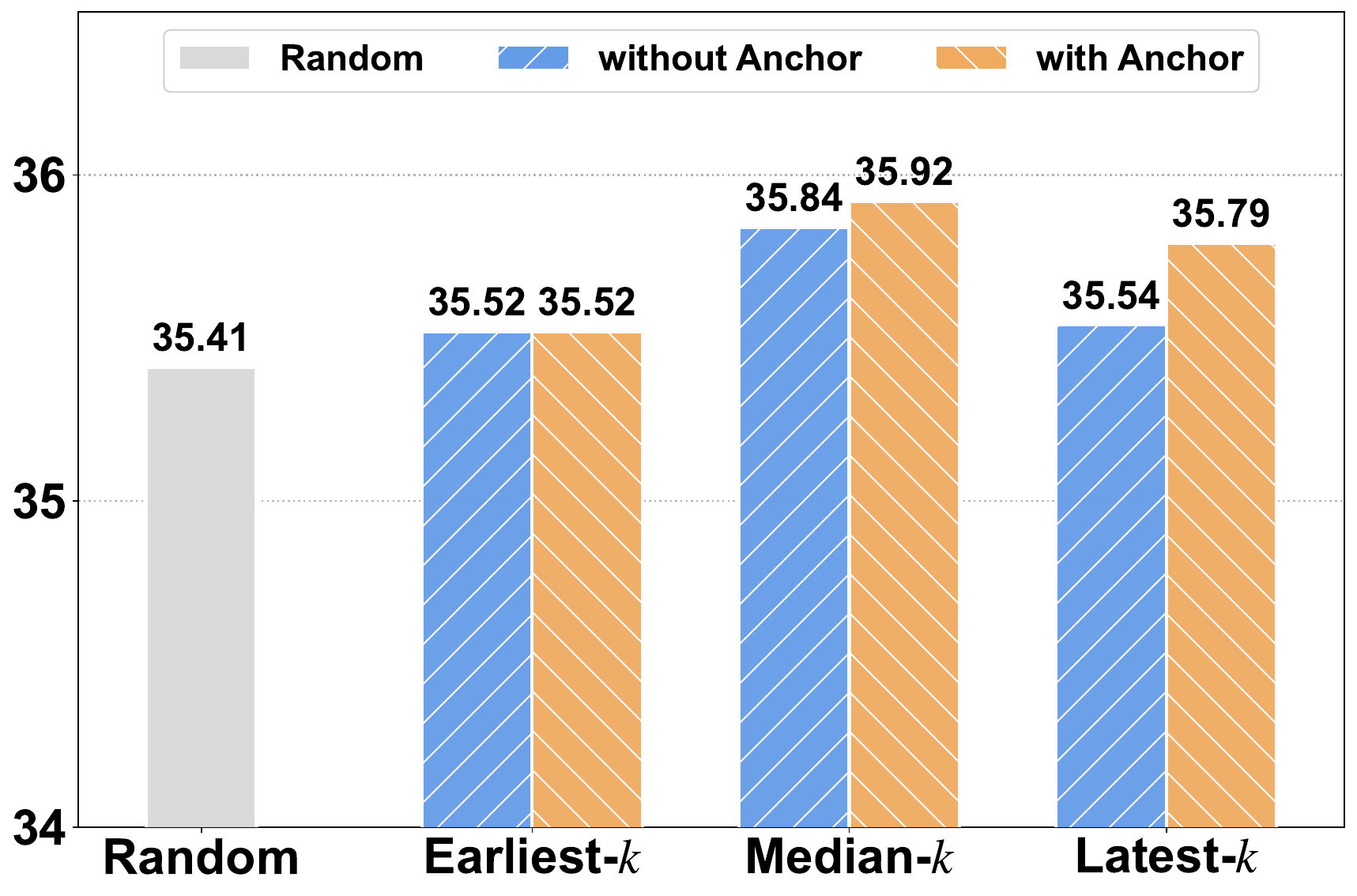}}
\caption{Effect of timestamp selection strategy ($K\!=\!6$, softmax aggregation). The middle + anchor (our default) achieves the best overall performance.}
\label{fig_rq3c}
\Description{fig_rq3c}
\end{figure*}

\begin{table*}[th]
\caption{Effect of timestamp selection strategy ($K\!=\!6$, softmax aggregation). Middle + anchor (our default) achieves the best overall performance. $\checkmark$/$\times$ indicates whether the earliest-year anchor is enforced. \textsuperscript{*}\footnotesize{Earliest selection naturally includes the earliest year; anchor is redundant.}}
\label{table_select}
\centering
\begin{tabular}{@{}lc|cccc|cccc|cccc@{}}
\toprule
\multirow{2}{*}{\textbf{Strategy}} & \multirow{2}{*}{\textbf{Anchor}} & \multicolumn{4}{c|}{\textbf{DB15K}} & \multicolumn{4}{c|}{\textbf{MKG-W}} & \multicolumn{4}{c}{\textbf{MKG-Y}} \\
                                    &        & MRR & H@1 & H@3 & H@10 & MRR & H@1 & H@3 & H@10 & MRR & H@1 & H@3 & H@10 \\ \midrule
Random & $\times$ & 38.38 & 30.86 & 41.93 & 52.82 & 38.50 & 32.24 & 40.95 & 49.88 & 39.01 & 35.41 & 40.59 & 45.14 \\
Earliest & $\checkmark$\textsuperscript{*} & 38.52 & 30.85 & 42.21 & 53.28 & 38.73 & 32.36 & 41.13 & 50.71 & 39.10 & 35.52 & 40.72 & 45.17 \\
Middle & $\times$ & 39.33 & 31.25 & 42.38 & 53.43 & 38.94 & 32.50 & \textbf{41.67} & 50.69 & 39.36 & 35.84 & 40.86 & 45.64 \\
\textbf{Middle} & $\checkmark$ & \textbf{39.75} & \textbf{31.76} & \textbf{42.96} & \textbf{53.86} & 38.99 & \textbf{32.67} & 41.38 & 50.46 & \textbf{39.62} & \textbf{35.92} & \textbf{41.62} & \textbf{46.58} \\
Latest & $\times$ & 38.49 & 31.02 & 41.95 & 52.88 & \textbf{39.08} & 32.66 & 41.47 & \textbf{51.20} & 39.19 & 35.54 & 40.71 & 45.72 \\
Latest & $\checkmark$ & 38.36 & 30.68 & 41.90 & 53.18 & 38.61 & 32.43 & 40.75 & 50.42 & 39.27 & 35.79 & 40.78 & 45.51 \\ \bottomrule
\end{tabular}
\end{table*}

We analyze three design dimensions of multi-timestamp modeling: the timestamp selection strategy, the number of selected timestamps~$K$, and the attention-based aggregation strategy.

\noindent \textbf{Effect of Selection Strategy.}
We compare six timestamp selection strategies, varying the selection window (earliest, middle, latest, random) and the presence of the earliest-year anchor constraint (Figure~\ref{fig4}); details are provided in Appendix Effect of Selection Strategy.
Figure~\ref{fig_rq3c} reports results with $K\!=\!6$ and softmax aggregation.
Middle + anchor (our default) achieves the best Hits@1 on MKG-W and DB15K and competitive results on MKG-Y, confirming that selecting timestamps around the median activity window with a historical anchor is the most robust strategy. We attribute the advantage of the middle window to its representativeness: the median sits at the center of an entity's activity span and is the least sensitive to outliers, whereas the earliest or latest years are often boundary events (e.g., an initial creation date or a much later re-release) that describe only one end of the entity's temporal extent. The earliest-year anchor then re-introduces the single most informative boundary, so the model retains a stable historical reference without letting an extreme value dominate the aggregated signal.
Three patterns are noteworthy.
First, the earliest-year anchor consistently improves median selection. On DB15K, adding the anchor increases Hits@1 by 1.01, indicating that the earliest timestamp provides a valuable historical reference that complements the central activity window.
Second, random selection performs worst across all datasets, confirming that the temporal ordering of timestamps carries meaningful information and should not be treated as an unstructured set.
Third, latest + anchor performs worse compared to latest alone on MKG-W ($-$0.23 Hits@1), suggesting that when the anchor is far from the selected cluster, the resulting bimodal distribution weakens the aggregated signal.

\noindent \textbf{Effect of Timestamp Count $K$.}
We fix the selection strategy to median-$K$ with the earliest anchor and vary $K$ from 1 to 10. Figure~\ref{fig_ksweep} reports the results on all three datasets. A clear inverted-U pattern emerges: performance improves as $K$ increases from~1, peaks around $K\!=\!4$-$6$, and declines for larger~$K$. The optimal~$K$ is dataset-dependent. MKG-W and DB15K peak at $K\!=\!6$, while MKG-Y peaks at $K\!=\!4$, but the general trend is highly consistent. At $K\!=\!1$, the model relies on a single timestamp and is fragile to extraction errors (e.g., DB15K Hits@1 drops by 1.52 compared to $K\!=\!6$). At $K\!\geq\!8$, noisy or incidental timestamps dilute the temporal signal. The range $K\!\in\![4,6]$ balances \emph{temporal specificity} (sufficient to be discriminative) with \emph{temporal coverage} (sufficient to ensure robustness), supporting the design of our attention pooling mechanism.

\begin{figure}[th]
\centering
\subfigure[MRR]{\includegraphics[width=0.49\linewidth]{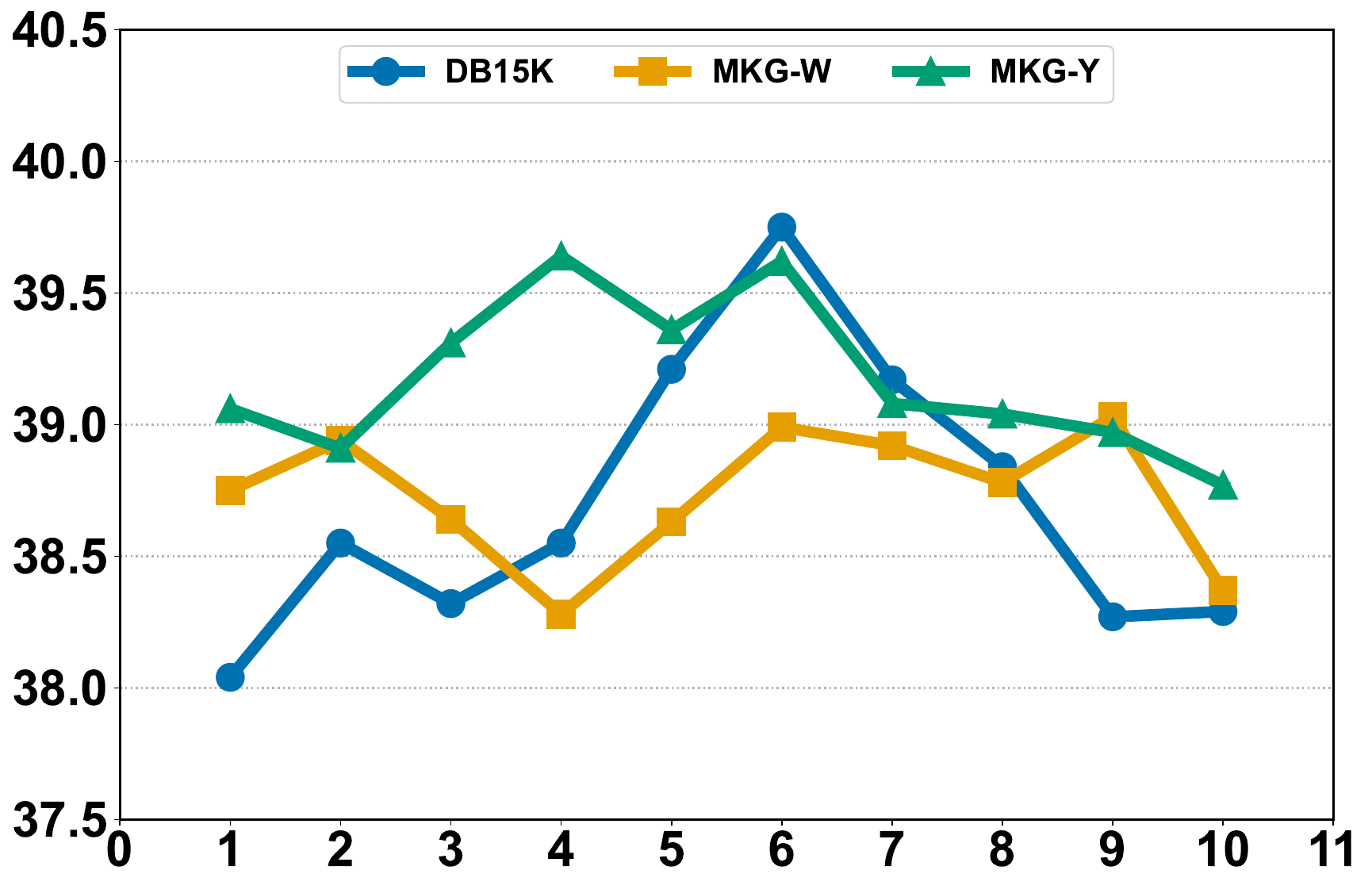}}
\subfigure[Hits@1]{\includegraphics[width=0.49\linewidth]{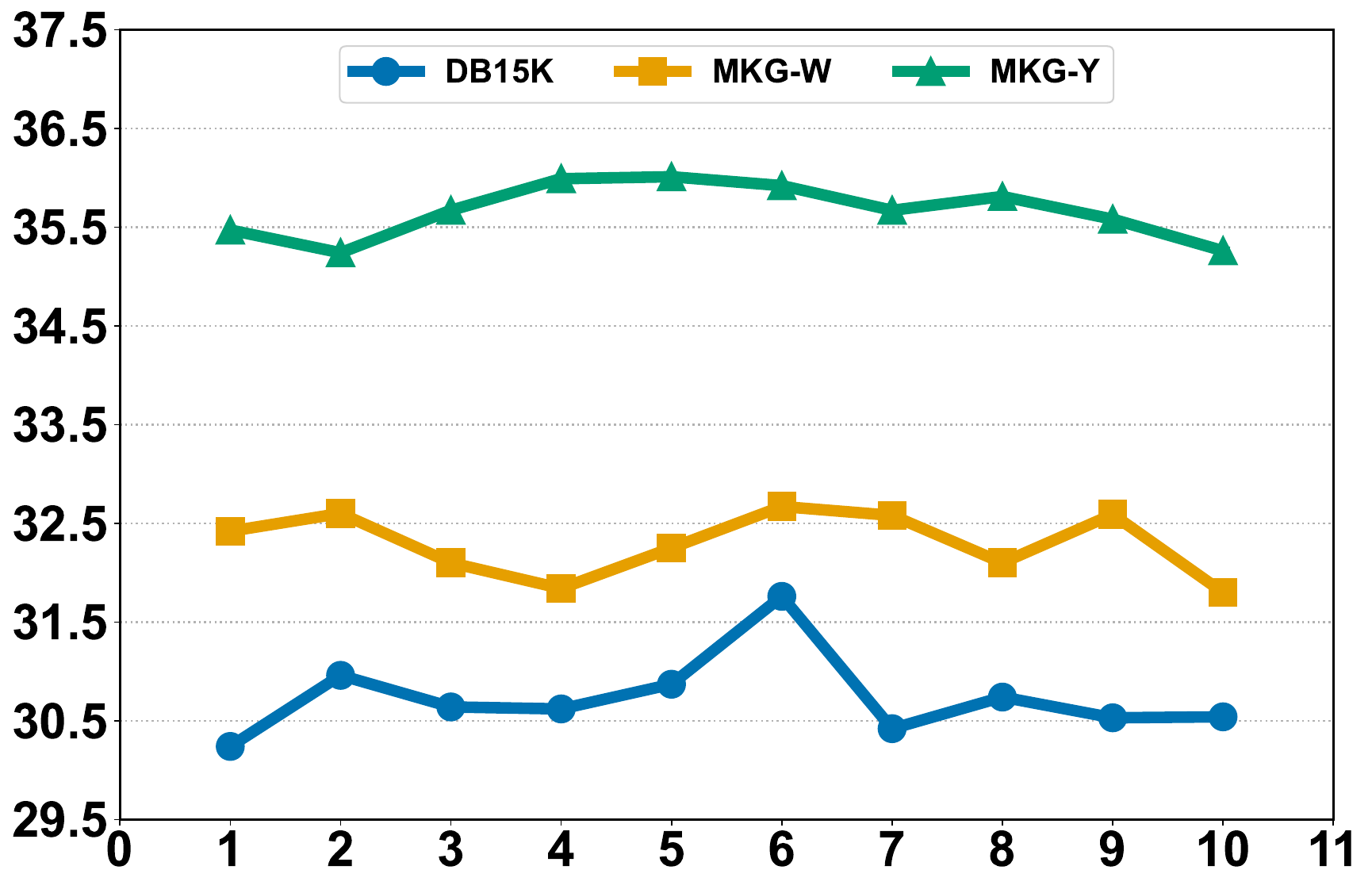}}\\
\subfigure[Hits@3]{\includegraphics[width=0.49\linewidth]{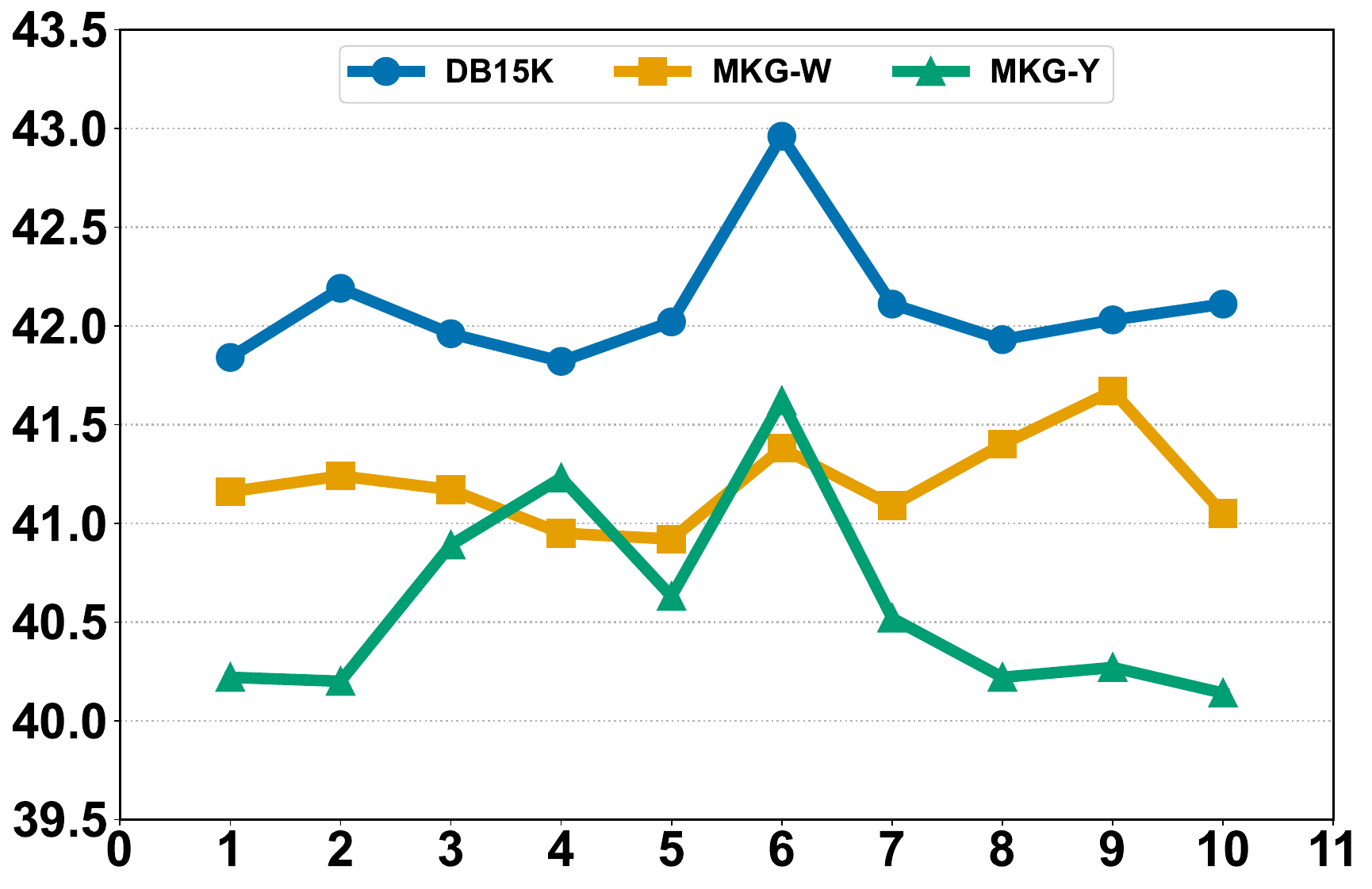}}
\subfigure[Hits@10]{\includegraphics[width=0.49\linewidth]{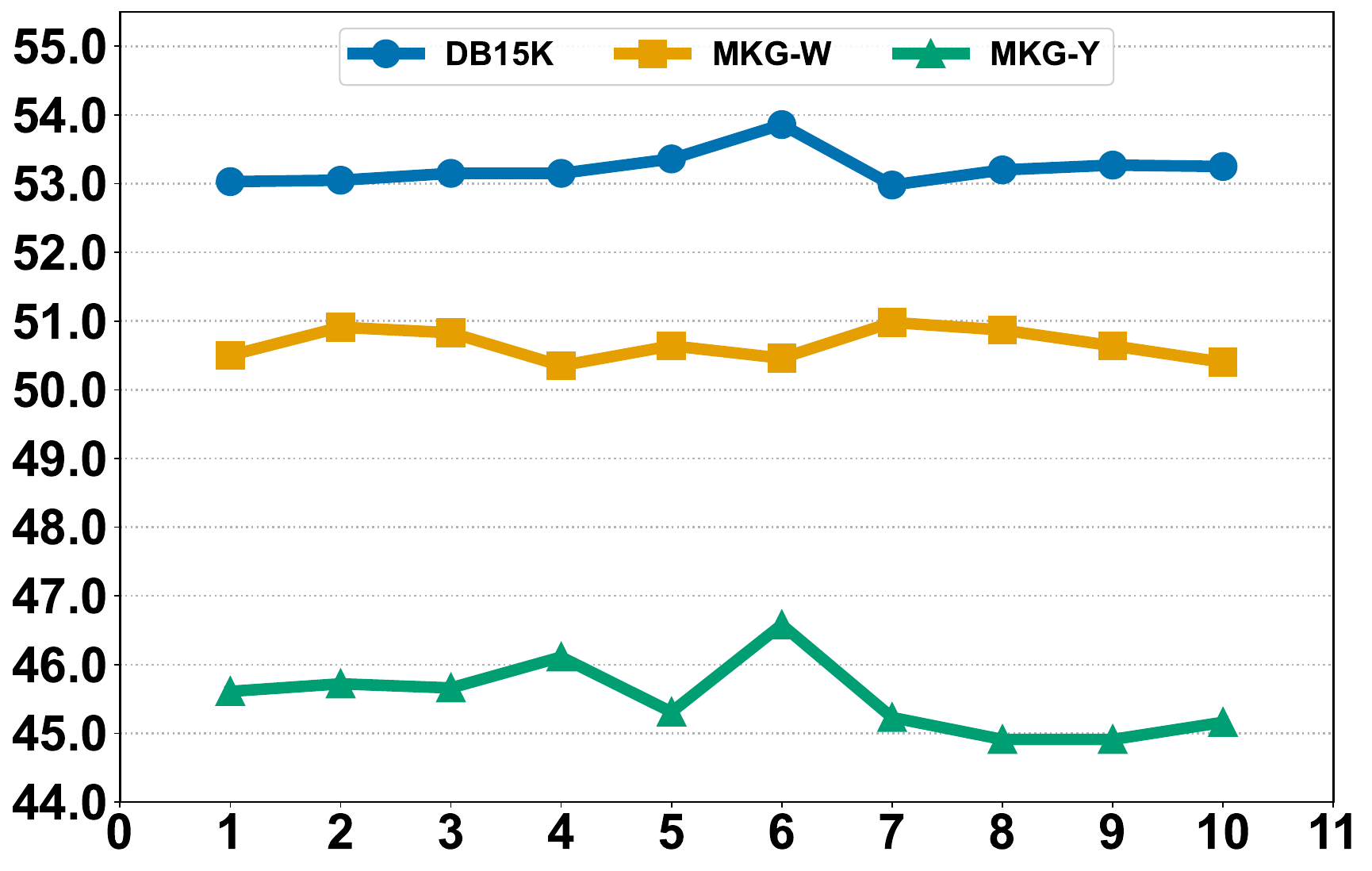}}
\caption{Effect of the number of selected timestamps~$K$ (median-$K$ selection with earliest anchor). Performance follows an inverted-U pattern, peaking at $K\!=\!4$-$6$.}
\label{fig_ksweep}
\Description{fig_ksweep}
\end{figure}

\noindent \textbf{Effect of Aggregation Strategy.}
We compare three attention normalization functions for timestamp aggregation with $K\!=\!6$ fixed: softmax (dense, where all timestamps receive nonzero weight), sparsemax (some timestamps receive exactly zero weight), and entmax~1.5 (an interpolation between the two).
Table~\ref{table_agg} reports the results.
Softmax achieves the best Hits@1 on all three datasets, while sparsemax shows a marginally higher MRR on MKG-W (+0.04).
Entmax~1.5 underperforms softmax on all three datasets, suggesting that overly aggressive sparsification truncates useful temporal evidence.
Overall, softmax with a well-tuned temperature ($\tau_a\!=\!0.5$-$0.6$) provides sufficient selectivity without hard sparsification, and we adopt it as the default.

\begin{table}[htb]
\caption{Effect of aggregation strategy on timestamp attention pooling ($K\!=\!6$). Softmax is the most stable across datasets.}
\label{table_agg}
\centering
\begin{tabular}{@{}cl|cccc@{}}
\toprule
& \textbf{Strategy} & \textbf{MRR} & \textbf{H@1} & \textbf{H@3} & \textbf{H@10} \\ \midrule
\multirow{3}{*}{\rotatebox{90}{\textbf{DB15K}}}
& Softmax     & \textbf{39.75} & \textbf{31.76} & \textbf{42.96} & \textbf{53.86} \\
& Sparsemax   & 38.50 & 30.95 & 42.04 & 53.09 \\
& Entmax 1.5  & 38.52 & 30.89 & 42.26 & 53.10 \\ \midrule
\multirow{3}{*}{\rotatebox{90}{\textbf{MKG-W}}}
& Softmax     & 38.99 & \textbf{32.67} & \textbf{41.38} & 50.46 \\
& Sparsemax   & \textbf{39.03} & \textbf{32.76} & 41.12 & \textbf{50.85} \\
& Entmax 1.5  & 38.86 & 32.50 & 41.26 & 50.82 \\ \midrule
\multirow{3}{*}{\rotatebox{90}{\textbf{MKG-Y}}}
& Softmax     & \textbf{39.62} & 35.92 & \textbf{41.62} & \textbf{46.58} \\
& Sparsemax   & 39.52 & \textbf{35.94} & 41.25 & 45.48 \\
& Entmax 1.5  & 38.94 & 35.34 & 40.33 & 45.57 \\ \bottomrule
\end{tabular}
\end{table}

\subsection{Robustness to Timestamp Noise (RQ4)}
\label{sec:rq4}

To assess sensitivity to timestamp quality, we corrupt entity timestamps at varying rates.
For each corruption rate $\rho \in \{20\%, 40\%, 60\%, 80\%, 100\%\}$, we randomly select a fraction $\rho$ of entities and replace all their timestamps with years uniformly sampled from the global year pool.
The clean setting ($\rho\!=\!0\%$) uses the original Full Time Imprint results.
All other hyper-parameters remain fixed.

\begin{figure}[htb]
\centering
\subfigure[MRR]{\includegraphics[width=0.49\linewidth]{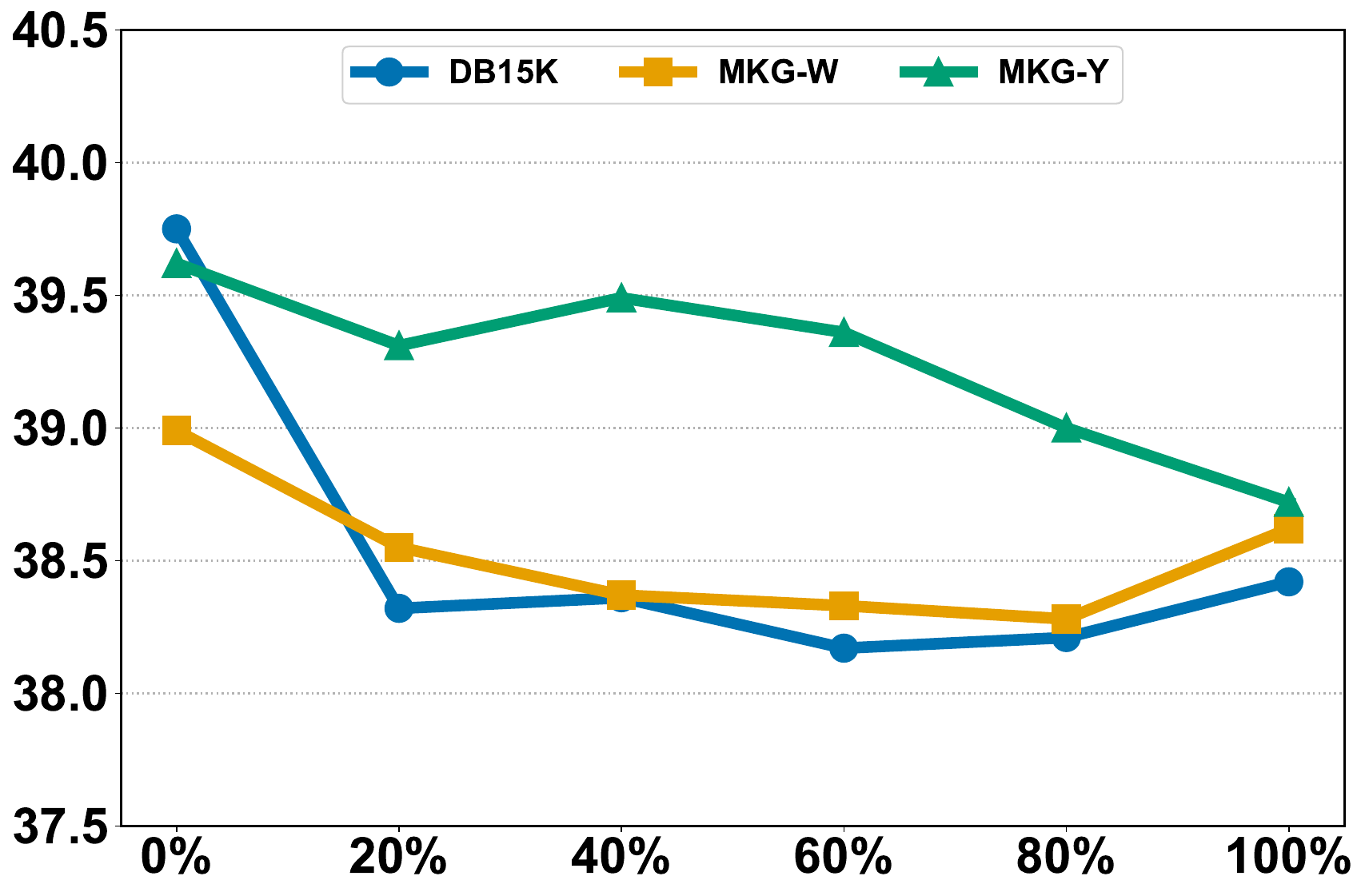}}
\subfigure[Hits@1]{\includegraphics[width=0.49\linewidth]{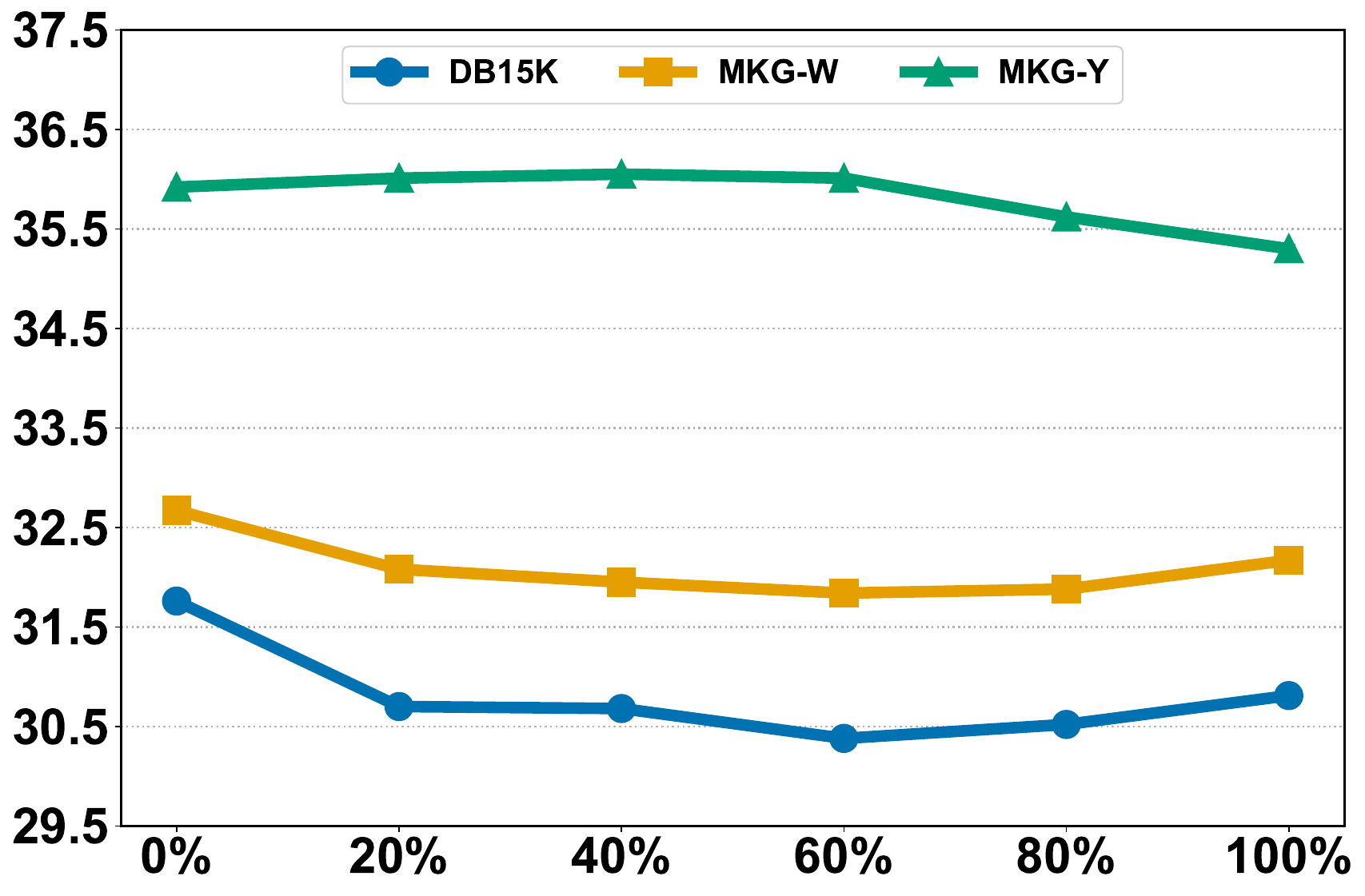}}\\
\subfigure[Hits@3]{\includegraphics[width=0.49\linewidth]{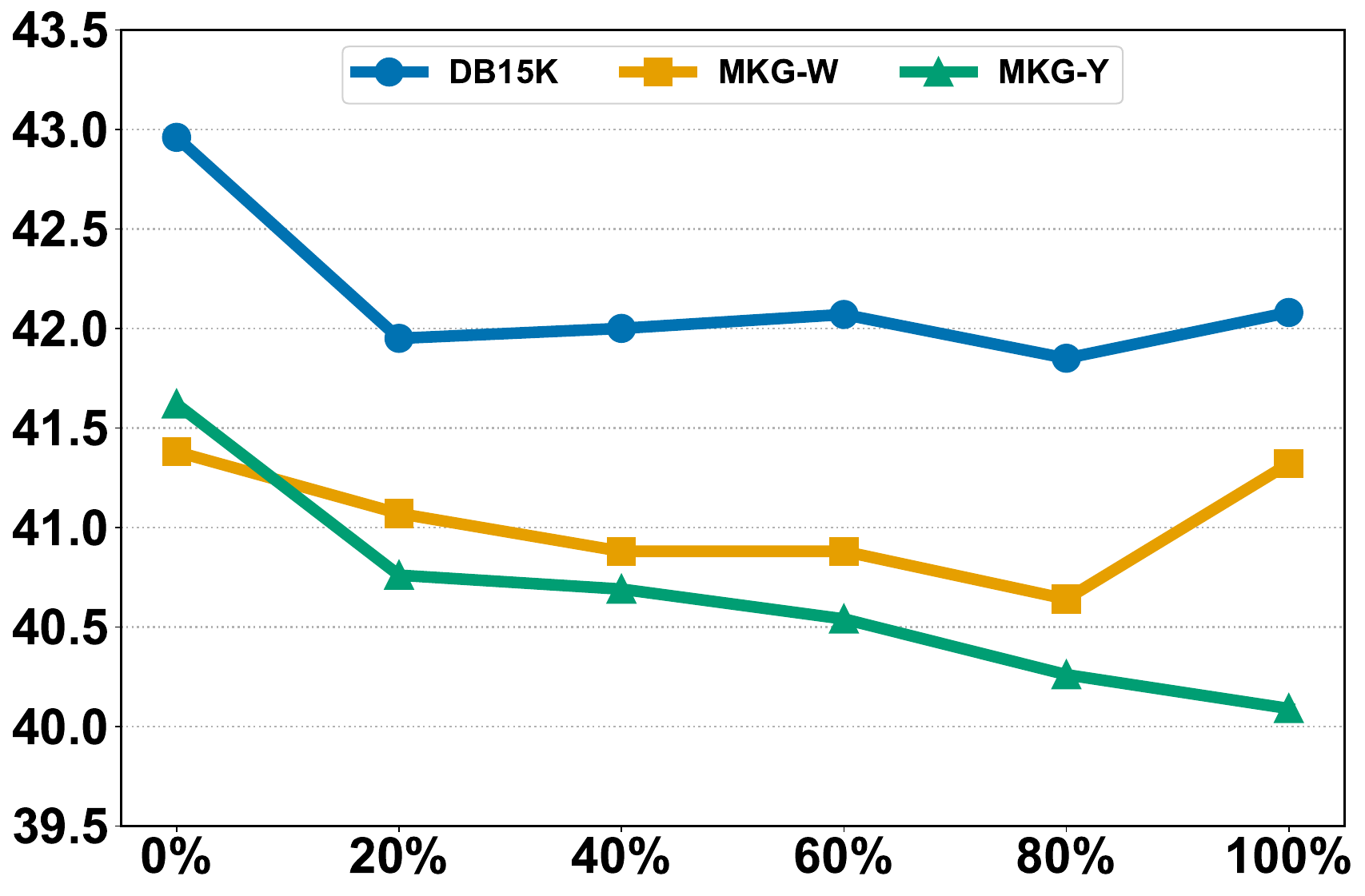}}
\subfigure[Hits@10]{\includegraphics[width=0.49\linewidth]{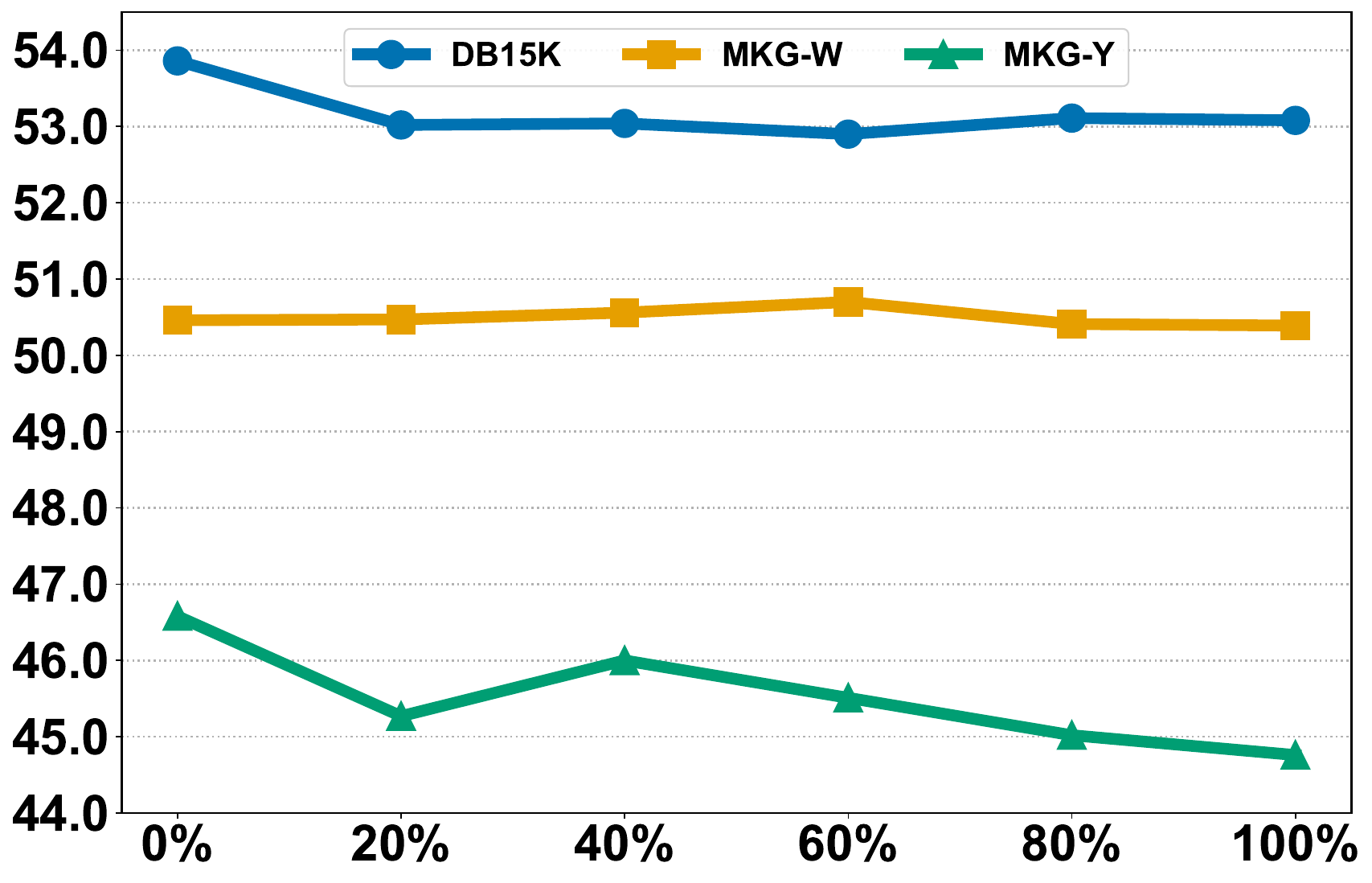}}
\caption{Performance under increasing timestamp corruption rates. Time Imprint degrades gracefully up to $\sim$40\% noise. The non-monotonic recovery at 100\% reflects the model reverting to a time-agnostic state when all timestamps are random.}
\label{fig_noise}
\Description{fig_noise}
\end{figure}

Figure~\ref{fig_noise} plots the Hits@1 degradation curves.
On MKG-W and DB15K, performance degrades gracefully up to 40\% corruption: MKG-W Hits@1 drops by only 0.72 (from 32.67 to 31.95), while DB15K drops by 1.08 (from 31.76 to 30.68).
This robustness can be attributed to the attention pooling mechanism, which downweights timestamps that deviate from the set centroid.

An interesting non-monotonic pattern appears at 100\% corruption: performance \emph{recovers} compared to 80\% on both MKG-W (32.17 vs.\ 31.88) and DB15K (30.81 vs.\ 30.52).
When all timestamps are random, their embeddings are distributed uniformly in feature space, causing attention weights to become nearly uniform. As a result, the model effectively ignores the temporal modality and reverts to a no-time baseline.
At 80\% corruption, the mixture of a few real timestamps with predominantly random ones creates the most adversarial condition, as the attention mechanism cannot reliably distinguish signal from noise.

MKG-Y exhibits a distinct pattern: Hits@1 at 20-60\% corruption (36.01-36.05) slightly exceeds the clean setting (35.92).
We do not claim a definitive cause. The gains are very small (within roughly 0.1 Hits@1) and may partly reflect run-to-run variance rather than a systematic effect. Beyond that, one plausible mechanism is mild regularization: if the original timestamps for some MKG-Y entities are themselves noisy or only weakly informative, replacing a fraction of them with random years could discourage the model from overfitting to spurious temporal patterns, analogous to input-level noise injection. We treat this as a hypothesis rather than a conclusion and leave a targeted verification, e.g., auditing the affected MKG-Y timestamps, to future work.
Beyond 60\% corruption, the expected degradation trend resumes.

\section{Conclusion}

We introduced \textbf{Time Imprint}, a framework that treats time as an entity-level modality and aligns temporal, textual, and visual representations through contrastive learning. Experiments on three MMKG benchmarks demonstrate state-of-the-art performance on MKG-W and MKG-Y, with improvements of up to 6.07\% in Hits@1 overall and 58\% on the top-1\% ambiguity subset. Our analyses confirm that modeling time as a modality is most beneficial when visual and textual features alone cannot disambiguate entities, and that careful timestamp selection is essential for suppressing noise.
\balance
\begin{acks}
The first author is supported by the China Scholarship Council (NO. 202206540007) and the University of Amsterdam. This funding source had no influence on the study design, data collection, analysis, or manuscript preparation and approval.
\end{acks}

\section*{Generative AI Disclosure}
We used a large language model (LLM) to assist with English proofreading and minor grammar edits in parts of the manuscript.

\section*{Ethics and Privacy Statement}
This work uses only publicly available benchmark datasets (DB15K, MKG-W, MKG-Y) about public entities, and involves no human subjects or personal data, so we foresee minimal direct risk. We release our code to support transparency and responsible use.

\clearpage
\balance
\bibliographystyle{ACM-Reference-Format}
\bibliography{sample-base}

@inproceedings{mmkgprototypesWebconf,
author = {Yang, Shundong and Yang, Jing and Jiang, Xiaowen and Gao, Yuan and Yang, Laurence T. and Luo, Ruikun and Yang, Jieming},
title = {Towards Multimodal Inductive Learning: Adaptively Embedding MMKG via Prototypes},
year = {2025},
isbn = {9798400712746},
publisher = {Association for Computing Machinery},
address = {New York, NY, USA},
booktitle = {Proceedings of the ACM on Web Conference 2025},
pages = {109–118},
numpages = {10},
location = {Sydney NSW, Australia},
series = {WWW '25}
}

@inproceedings{blp2021,
author = {Daza, Daniel and Cochez, Michael and Groth, Paul},
title = {Inductive Entity Representations from Text via Link Prediction},
year = {2021},
isbn = {9781450383127},
publisher = {Association for Computing Machinery},
address = {New York, NY, USA},
booktitle = {Proceedings of the Web Conference 2021},
pages = {798–808},
numpages = {11},
location = {Ljubljana, Slovenia},
series = {WWW '21}
}

@inproceedings{10.5555/3172077.3172327,
author = {Xie, Ruobing and Liu, Zhiyuan and Luan, Huanbo and Sun, Maosong},
title = {Image-embodied knowledge representation learning},
year = {2017},
isbn = {9780999241103},
publisher = {AAAI Press},
booktitle = {Proceedings of the 26th International Joint Conference on Artificial Intelligence},
pages = {3140–3146},
numpages = {7},
location = {Melbourne, Australia},
series = {IJCAI'17}
}

@inproceedings{lin-etal-2022-multi,
    title = "Multi-modal Contrastive Representation Learning for Entity Alignment",
    author = "Lin, Zhenxi  and
      Zhang, Ziheng  and
      Wang, Meng  and
      Shi, Yinghui  and
      Wu, Xian  and
      Zheng, Yefeng",
    editor = "Calzolari, Nicoletta  and
      Huang, Chu-Ren  and
      Kim, Hansaem  and
      Pustejovsky, James  and
      Wanner, Leo  and
      Choi, Key-Sun  and
      Ryu, Pum-Mo  and
      Chen, Hsin-Hsi  and
      Donatelli, Lucia  and
      Ji, Heng  and
      Kurohashi, Sadao  and
      Paggio, Patrizia  and
      Xue, Nianwen  and
      Kim, Seokhwan  and
      Hahm, Younggyun  and
      He, Zhong  and
      Lee, Tony Kyungil  and
      Santus, Enrico  and
      Bond, Francis  and
      Na, Seung-Hoon",
    booktitle = "Proceedings of the 29th International Conference on Computational Linguistics",
    month = oct,
    year = "2022",
    address = "Gyeongju, Republic of Korea",
    publisher = "International Committee on Computational Linguistics",
    pages = "2572--2584"
}

@inproceedings{NEURIPS2022_ffdb280e,
 author = {Cao, Zongsheng and Xu, Qianqian and Yang, Zhiyong and He, Yuan and Cao, Xiaochun and Huang, Qingming},
 booktitle = {Advances in Neural Information Processing Systems},
 editor = {S. Koyejo and S. Mohamed and A. Agarwal and D. Belgrave and K. Cho and A. Oh},
 pages = {39090--39102},
 publisher = {Curran Associates, Inc.},
 title = {OTKGE: Multi-modal Knowledge Graph Embeddings via Optimal Transport},
 volume = {35},
 year = {2022}
}

@inproceedings{zhao-etal-2022-mose,
    title = "{M}o{SE}: Modality Split and Ensemble for Multimodal Knowledge Graph Completion",
    author = "Zhao, Yu  and
      Cai, Xiangrui  and
      Wu, Yike  and
      Zhang, Haiwei  and
      Zhang, Ying  and
      Zhao, Guoqing  and
      Jiang, Ning",
    editor = "Goldberg, Yoav  and
      Kozareva, Zornitsa  and
      Zhang, Yue",
    booktitle = "Proceedings of the 2022 Conference on Empirical Methods in Natural Language Processing",
    month = dec,
    year = "2022",
    address = "Abu Dhabi, United Arab Emirates",
    publisher = "Association for Computational Linguistics",
    pages = "10527--10536"
}

@inproceedings{zhang-etal-2024-unleashing,
    title = "Unleashing the Power of Imbalanced Modality Information for Multi-modal Knowledge Graph Completion",
    author = "Zhang, Yichi  and
      Chen, Zhuo  and
      Liang, Lei  and
      Chen, Huajun  and
      Zhang, Wen",
    editor = "Calzolari, Nicoletta  and
      Kan, Min-Yen  and
      Hoste, Veronique  and
      Lenci, Alessandro  and
      Sakti, Sakriani  and
      Xue, Nianwen",
    booktitle = "Proceedings of the 2024 Joint International Conference on Computational Linguistics, Language Resources and Evaluation (LREC-COLING 2024)",
    month = may,
    year = "2024",
    address = "Torino, Italia",
    publisher = "ELRA and ICCL",
    pages = "17120--17130"
}

@article{Shang_Zhao_Liu_Wang_2024, title={LAFA: Multimodal Knowledge Graph Completion with Link Aware Fusion and Aggregation}, volume={38}, abstractNote={Recently, an enormous amount of research has emerged on multimodal knowledge graph completion (MKGC), which seeks to extract knowledge from multimodal data and predict the most plausible missing facts to complete a given multimodal knowledge graph (MKG). However, existing MKGC approaches largely ignore that visual information may introduce noise and lead to uncertainty when adding them to the traditional KG embeddings due to the contribution of each associated image to entity is different in diverse link scenarios. Moreover, treating each triple independently when learning entity embeddings leads to local structural and the whole graph information missing. To address these challenges, we propose a novel link aware fusion and aggregation based multimodal knowledge graph completion model named LAFA, which is composed of link aware fusion module and link aware aggregation module. The link aware fusion module alleviates noise of irrelevant visual information by calculating the importance between an entity and its associated images in different link scenarios, and fuses the visual and structural embeddings according to the importance through our proposed modality embedding fusion mechanism. The link aware aggregation module assigns neighbor structural information to a given central entity by calculating the importance between the entity and its neighbors, and aggregating the fused embeddings through linear combination according to the importance. Extensive experiments on standard datasets validate that LAFA can obtain state-of-the-art performance.}, number={8}, journal={Proceedings of the AAAI Conference on Artificial Intelligence}, author={Shang, Bin and Zhao, Yinliang and Liu, Jun and Wang, Di}, year={2024}, month={Mar.}, pages={8957-8965} }

@article{Liu_Chen_Roth_Collier_2021, title={Visual Pivoting for (Unsupervised) Entity Alignment}, volume={35}, abstractNote={This work studies the use of visual semantic representations to align entities in heterogeneous knowledge graphs (KGs). Images are natural components of many existing KGs. By combining visual knowledge with other auxiliary information, we show that the proposed new approach, EVA, creates a holistic entity representation that provides strong signals for cross-graph entity alignment. Besides, previous entity alignment methods require human labelled seed alignment, restricting availability. EVA provides a completely unsupervised solution by leveraging the visual similarity of entities to create an initial seed dictionary (visual pivots). Experiments on benchmark data sets DBP15k and DWY15k show that EVA offers state-of-the-art performance on both monolingual and cross-lingual entity alignment tasks. Furthermore, we discover that images are particularly useful to align long-tail KG entities, which inherently lack the structural contexts necessary for capturing the correspondences. Code release: https://github.com/cambridgeltl/eva; project page: http://cogcomp.org/page/publication view/927.}, number={5}, journal={Proceedings of the AAAI Conference on Artificial Intelligence}, author={Liu, Fangyu and Chen, Muhao and Roth, Dan and Collier, Nigel}, year={2021}, month={May}, pages={4257-4266} }

@article{GUO2021598,
title = {Multi-modal entity alignment in hyperbolic space},
journal = {Neurocomputing},
volume = {461},
pages = {598-607},
year = {2021},
issn = {0925-2312},
author = {Hao Guo and Jiuyang Tang and Weixin Zeng and Xiang Zhao and Li Liu}
}

@inproceedings{jiang-etal-2016-towards,
    title = "Towards Time-Aware Knowledge Graph Completion",
    author = "Jiang, Tingsong  and
      Liu, Tianyu  and
      Ge, Tao  and
      Sha, Lei  and
      Chang, Baobao  and
      Li, Sujian  and
      Sui, Zhifang",
    editor = "Matsumoto, Yuji  and
      Prasad, Rashmi",
    booktitle = "Proceedings of {COLING} 2016, the 26th International Conference on Computational Linguistics: Technical Papers",
    month = dec,
    year = "2016",
    address = "Osaka, Japan",
    publisher = "The COLING 2016 Organizing Committee",
    pages = "1715--1724"
}

@article{Goel_Kazemi_Brubaker_Poupart_2020, title={Diachronic Embedding for Temporal Knowledge Graph Completion}, volume={34}, abstractNote={&lt;p&gt;Knowledge graphs (KGs) typically contain temporal facts indicating relationships among entities at different times. Due to their incompleteness, several approaches have been proposed to infer new facts for a KG based on the existing ones–a problem known as &lt;em&gt;KG completion&lt;/em&gt;. KG embedding approaches have proved effective for KG completion, however, they have been developed mostly for static KGs. Developing temporal KG embedding models is an increasingly important problem. In this paper, we build novel models for temporal KG completion through equipping static models with a diachronic entity embedding function which provides the characteristics of entities at &lt;em&gt;any&lt;/em&gt; point in time. This is in contrast to the existing temporal KG embedding approaches where only static entity features are provided. The proposed embedding function is model-agnostic and can be potentially combined with any static model. We prove that combining it with SimplE, a recent model for static KG embedding, results in a fully expressive model for temporal KG completion. Our experiments indicate the superiority of our proposal compared to existing baselines.&lt;/p&gt;}, number={04}, journal={Proceedings of the AAAI Conference on Artificial Intelligence}, author={Goel, Rishab and Kazemi, Seyed Mehran and Brubaker, Marcus and Poupart, Pascal}, year={2020}, month={Apr.}, pages={3988-3995} }

@inproceedings{jin-etal-2020-recurrent,
    title = "Recurrent Event Network: Autoregressive Structure Inferenceover Temporal Knowledge Graphs",
    author = "Jin, Woojeong  and
      Qu, Meng  and
      Jin, Xisen  and
      Ren, Xiang",
    editor = "Webber, Bonnie  and
      Cohn, Trevor  and
      He, Yulan  and
      Liu, Yang",
    booktitle = "Proceedings of the 2020 Conference on Empirical Methods in Natural Language Processing (EMNLP)",
    month = nov,
    year = "2020",
    address = "Online",
    publisher = "Association for Computational Linguistics",
    pages = "6669--6683"
}

@article{Zhu_Chen_Fan_Cheng_Zhang_2021, title={Learning from History: Modeling Temporal Knowledge Graphs with Sequential Copy-Generation Networks}, volume={35}, abstractNote={Large knowledge graphs often grow to store temporal facts that model the dynamic relations or interactions of entities along the timeline. Since such temporal knowledge graphs often suffer from incompleteness, it is important to develop time-aware representation learning models that help to infer the missing temporal facts. While the temporal facts are typically evolving, it is observed that many facts often show a repeated pattern along the timeline, such as economic crises and diplomatic activities. This observation indicates that a model could potentially learn much from the known facts appeared in history. To this end, we propose a new representation learning model for temporal knowledge graphs, namely CyGNet, based on a novel time-aware copy-generation mechanism. CyGNet is not only able to predict future facts from the whole entity vocabulary, but also capable of identifying facts with repetition and accordingly predicting such future facts with reference to the known facts in the past. We evaluate the proposed method on the knowledge graph completion task using five benchmark datasets. Extensive experiments demonstrate the effectiveness of CyGNet for predicting future facts with repetition as well as de novo fact prediction.}, number={5}, journal={Proceedings of the AAAI Conference on Artificial Intelligence}, author={Zhu, Cunchao and Chen, Muhao and Fan, Changjun and Cheng, Guangquan and Zhang, Yan}, year={2021}, month={May}, pages={4732-4740} }

@inproceedings{dasgupta-etal-2018-hyte,
    title = "{H}y{TE}: Hyperplane-based Temporally aware Knowledge Graph Embedding",
    author = "Dasgupta, Shib Sankar  and
      Ray, Swayambhu Nath  and
      Talukdar, Partha",
    editor = "Riloff, Ellen  and
      Chiang, David  and
      Hockenmaier, Julia  and
      Tsujii, Jun{'}ichi",
    booktitle = "Proceedings of the 2018 Conference on Empirical Methods in Natural Language Processing",
    month = oct # "-" # nov,
    year = "2018",
    address = "Brussels, Belgium",
    publisher = "Association for Computational Linguistics",
    pages = "2001--2011"
}

@inproceedings{10.1145/3404835.3462963,
author = {Li, Zixuan and Jin, Xiaolong and Li, Wei and Guan, Saiping and Guo, Jiafeng and Shen, Huawei and Wang, Yuanzhuo and Cheng, Xueqi},
title = {Temporal Knowledge Graph Reasoning Based on Evolutional Representation Learning},
year = {2021},
isbn = {9781450380379},
publisher = {Association for Computing Machinery},
address = {New York, NY, USA},
booktitle = {Proceedings of the 44th International ACM SIGIR Conference on Research and Development in Information Retrieval},
pages = {408–417},
numpages = {10},
location = {Virtual Event, Canada},
series = {SIGIR '21}
}

@inproceedings{zhang-etal-2024-lgre,
  title={Learning Granularity Representation for Temporal Knowledge Graph Completion},
  author = {Jinchuan Zhang and Tianqi Wan and Chong Mu and Guangxi Lu and Ling Tian},
  booktitle={Neural Information Processing: 31th International Conference, ICONIP 2024, Auckland, New Zealand, December 2--6, 2024},
  year={2024},
  organization={Springer}
}

@misc{wilcke2023endtoendlearningmultimodalknowledge,
      title={End-to-End Learning on Multimodal Knowledge Graphs}, 
      author={W. X. Wilcke and P. Bloem and V. de Boer and R. H. van t Veer},
      year={2023},
      eprint={2309.01169},
      archivePrefix={arXiv},
      primaryClass={cs.LG},
}

@inproceedings{10.1145/3626772.3657838,
author = {Zhao, Yu and Zhang, Ying and Zhou, Baohang and Qian, Xinying and Song, Kehui and Cai, Xiangrui},
title = {Contrast then Memorize: Semantic Neighbor Retrieval-Enhanced Inductive Multimodal Knowledge Graph Completion},
year = {2024},
isbn = {9798400704314},
publisher = {Association for Computing Machinery},
address = {New York, NY, USA},
booktitle = {Proceedings of the 47th International ACM SIGIR Conference on Research and Development in Information Retrieval},
pages = {102–111},
numpages = {10},
location = {Washington DC, USA},
series = {SIGIR '24}
}

@inproceedings{zhang-etal-2024-multi,
    title = "Multi-modal Semantic Understanding with Contrastive Cross-modal Feature Alignment",
    author = "Zhang, Ming  and
      Chang, Ke  and
      Wu, Yunfang",
    editor = "Calzolari, Nicoletta  and
      Kan, Min-Yen  and
      Hoste, Veronique  and
      Lenci, Alessandro  and
      Sakti, Sakriani  and
      Xue, Nianwen",
    booktitle = "Proceedings of the 2024 Joint International Conference on Computational Linguistics, Language Resources and Evaluation (LREC-COLING 2024)",
    month = may,
    year = "2024",
    address = "Torino, Italia",
    publisher = "ELRA and ICCL",
    pages = "11934--11943"
}

@inproceedings{huang-etal-2024-progressively,
    title = "Progressively Modality Freezing for Multi-Modal Entity Alignment",
    author = "Huang, Yani  and
      Zhang, Xuefeng  and
      Zhang, Richong  and
      Chen, Junfan  and
      Kim, Jaein",
    editor = "Ku, Lun-Wei  and
      Martins, Andre  and
      Srikumar, Vivek",
    booktitle = "Proceedings of the 62nd Annual Meeting of the Association for Computational Linguistics (Volume 1: Long Papers)",
    month = aug,
    year = "2024",
    address = "Bangkok, Thailand",
    publisher = "Association for Computational Linguistics",
    pages = "3477--3489"
}

@inproceedings{lin-etal-2023-techs,
    title = "{TECHS}: Temporal Logical Graph Networks for Explainable Extrapolation Reasoning",
    author = "Lin, Qika  and
      Liu, Jun  and
      Mao, Rui  and
      Xu, Fangzhi  and
      Cambria, Erik",
    editor = "Rogers, Anna  and
      Boyd-Graber, Jordan  and
      Okazaki, Naoaki",
    booktitle = "Proceedings of the 61st Annual Meeting of the Association for Computational Linguistics (Volume 1: Long Papers)",
    month = jul,
    year = "2023",
    address = "Toronto, Canada",
    publisher = "Association for Computational Linguistics",
    pages = "1281--1293"
}

@misc{cai2024surveytemporalknowledgegraph,
      title={A Survey on Temporal Knowledge Graph: Representation Learning and Applications}, 
      author={Li Cai and Xin Mao and Yuhao Zhou and Zhaoguang Long and Changxu Wu and Man Lan},
      year={2024},
      eprint={2403.04782},
      archivePrefix={arXiv},
      primaryClass={cs.CL},
}

@misc{ma2025transformerbasedmultimodalknowledgegraph,
      title={Transformer-Based Multimodal Knowledge Graph Completion with Link-Aware Contexts}, 
      author={Haodi Ma and Dzmitry Kasinets and Daisy Zhe Wang},
      year={2025},
      eprint={2501.15688},
      archivePrefix={arXiv},
      primaryClass={cs.CL},
}

@misc{peng2022beitv2maskedimage,
      title={BEiT v2: Masked Image Modeling with Vector-Quantized Visual Tokenizers}, 
      author={Zhiliang Peng and Li Dong and Hangbo Bao and Qixiang Ye and Furu Wei},
      year={2022},
      eprint={2208.06366},
      archivePrefix={arXiv},
      primaryClass={cs.CV},
}

@inproceedings{devlin-etal-2019-bert,
    title = "{BERT}: Pre-training of Deep Bidirectional Transformers for Language Understanding",
    author = "Devlin, Jacob  and
      Chang, Ming-Wei  and
      Lee, Kenton  and
      Toutanova, Kristina",
    editor = "Burstein, Jill  and
      Doran, Christy  and
      Solorio, Thamar",
    booktitle = "Proceedings of the 2019 Conference of the North {A}merican Chapter of the Association for Computational Linguistics: Human Language Technologies, Volume 1 (Long and Short Papers)",
    month = jun,
    year = "2019",
    address = "Minneapolis, Minnesota",
    publisher = "Association for Computational Linguistics",
    pages = "4171--4186"
}

@inproceedings{balazevic-etal-2019-tucker,
    title = "{T}uck{ER}: Tensor Factorization for Knowledge Graph Completion",
    author = "Balazevic, Ivana  and
      Allen, Carl  and
      Hospedales, Timothy",
    editor = "Inui, Kentaro  and
      Jiang, Jing  and
      Ng, Vincent  and
      Wan, Xiaojun",
    booktitle = "Proceedings of the 2019 Conference on Empirical Methods in Natural Language Processing and the 9th International Joint Conference on Natural Language Processing (EMNLP-IJCNLP)",
    month = nov,
    year = "2019",
    address = "Hong Kong, China",
    publisher = "Association for Computational Linguistics",
    pages = "5185--5194"
}

@inproceedings{DBLP:conf/aaai/ZhangCGXHLZC25,
  author       = {Yichi Zhang and
                  Zhuo Chen and
                  Lingbing Guo and
                  Yajing Xu and
                  Binbin Hu and
                  Ziqi Liu and
                  Wen Zhang and
                  Huajun Chen},
  title        = {Tokenization, Fusion, and Augmentation: Towards Fine-grained Multi-modal
                  Entity Representation},
  booktitle    = {{AAAI}},
  pages        = {13322--13330},
  publisher    = {{AAAI} Press},
  year         = {2025}
}

@inproceedings{liu2019mmkg,
author = {Liu, Ye and Li, Hui and Garcia-Duran, Alberto and Niepert, Mathias and Onoro-Rubio, Daniel and Rosenblum, David S.},
title = {MMKG: Multi-modal Knowledge Graphs},
year = {2019},
isbn = {978-3-030-21347-3},
publisher = {Springer-Verlag},
address = {Berlin, Heidelberg},
booktitle = {The Semantic Web: 16th International Conference, ESWC 2019, Portoro\v{z}, Slovenia, June 2–6, 2019, Proceedings},
pages = {459–474},
numpages = {16},
location = {Portoroz, Slovenia}
}

@inproceedings{10.1145/3503161.3548388,
author = {Xu, Derong and Xu, Tong and Wu, Shiwei and Zhou, Jingbo and Chen, Enhong},
title = {Relation-enhanced Negative Sampling for Multimodal Knowledge Graph Completion},
year = {2022},
isbn = {9781450392037},
publisher = {Association for Computing Machinery},
address = {New York, NY, USA},
booktitle = {Proceedings of the 30th ACM International Conference on Multimedia},
pages = {3857–3866},
numpages = {10},
location = {Lisboa, Portugal},
series = {MM '22}
}

@inproceedings{10.1145/3627673.3679545,
author = {Pahuja, Vardaan and Luo, Weidi and Gu, Yu and Tu, Cheng-Hao and Chen, Hong-You and Berger-Wolf, Tanya and Stewart, Charles and Gao, Song and Chao, Wei-Lun and Su, Yu},
title = {Reviving the Context: Camera Trap Species Classification as Link Prediction on Multimodal Knowledge Graphs},
year = {2024},
isbn = {9798400704369},
publisher = {Association for Computing Machinery},
address = {New York, NY, USA},
booktitle = {Proceedings of the 33rd ACM International Conference on Information and Knowledge Management},
pages = {1825–1835},
numpages = {11},
location = {Boise, ID, USA},
series = {CIKM '24}
}

@article{10.1145/3656579,
author = {Liang, Wanying and Meo, Pasquale De and Tang, Yong and Zhu, Jia},
title = {A Survey of Multi-modal Knowledge Graphs: Technologies and Trends},
year = {2024},
issue_date = {November 2024},
publisher = {Association for Computing Machinery},
address = {New York, NY, USA},
volume = {56},
number = {11},
issn = {0360-0300},
journal = {ACM Comput. Surv.},
month = jun,
articleno = {273},
numpages = {41}
}

@inproceedings{10.1145/3626772.3657800,
author = {Zhang, Yichi and Chen, Zhuo and Guo, Lingbing and Xu, Yajing and Hu, Binbin and Liu, Ziqi and Zhang, Wen and Chen, Huajun},
title = {NativE: Multi-modal Knowledge Graph Completion in the Wild},
year = {2024},
isbn = {9798400704314},
publisher = {Association for Computing Machinery},
address = {New York, NY, USA},
booktitle = {Proceedings of the 47th International ACM SIGIR Conference on Research and Development in Information Retrieval},
pages = {91–101},
numpages = {11},
location = {Washington DC, USA},
series = {SIGIR '24}
}

@inproceedings{lee-etal-2023-vista,
    title = "{VISTA}: Visual-Textual Knowledge Graph Representation Learning",
    author = "Lee, Jaejun  and
      Chung, Chanyoung  and
      Lee, Hochang  and
      Jo, Sungho  and
      Whang, Joyce",
    editor = "Bouamor, Houda  and
      Pino, Juan  and
      Bali, Kalika",
    booktitle = "Findings of the Association for Computational Linguistics: EMNLP 2023",
    month = dec,
    year = "2023",
    address = "Singapore",
    publisher = "Association for Computational Linguistics",
    pages = "7314--7328"
}

@inproceedings{10.1145/3543507.3583554,
author = {Li, Xinhang and Zhao, Xiangyu and Xu, Jiaxing and Zhang, Yong and Xing, Chunxiao},
title = {IMF: Interactive Multimodal Fusion Model for Link Prediction},
year = {2023},
isbn = {9781450394161},
publisher = {Association for Computing Machinery},
address = {New York, NY, USA},
booktitle = {Proceedings of the ACM Web Conference 2023},
pages = {2572–2580},
numpages = {9},
location = {Austin, TX, USA},
series = {WWW '23}
}

@InProceedings{10.1007/978-3-031-44693-1_10,
author="Zhang, Yichi
and Chen, Zhuo
and Zhang, Wen",
editor="Liu, Fei
and Duan, Nan
and Xu, Qingting
and Hong, Yu",
title="MACO: A Modality Adversarial and Contrastive Framework for Modality-Missing Multi-modal Knowledge Graph Completion",
booktitle="Natural Language Processing and Chinese Computing",
year="2023",
publisher="Springer Nature Switzerland",
address="Cham",
pages="123--134",
isbn="978-3-031-44693-1"
}

@inproceedings{NIPS2013_1cecc7a7,
 author = {Bordes, Antoine and Usunier, Nicolas and Garcia-Duran, Alberto and Weston, Jason and Yakhnenko, Oksana},
 booktitle = {Advances in Neural Information Processing Systems},
 editor = {C.J. Burges and L. Bottou and M. Welling and Z. Ghahramani and K.Q. Weinberger},
 pages = {},
 publisher = {Curran Associates, Inc.},
 title = {Translating Embeddings for Modeling Multi-relational Data},
 volume = {26},
 year = {2013}
}

@misc{yang2015embeddingentitiesrelationslearning,
      title={Embedding Entities and Relations for Learning and Inference in Knowledge Bases}, 
      author={Bishan Yang and Wen-tau Yih and Xiaodong He and Jianfeng Gao and Li Deng},
      year={2015},
      eprint={1412.6575},
      archivePrefix={arXiv},
      primaryClass={cs.CL},
}

@InProceedings{pmlr-v48-trouillon16,
  title = 	 {Complex Embeddings for Simple Link Prediction},
  author = 	 {Trouillon, Théo and Welbl, Johannes and Riedel, Sebastian and Gaussier, Eric and Bouchard, Guillaume},
  booktitle = 	 {Proceedings of The 33rd International Conference on Machine Learning},
  pages = 	 {2071--2080},
  year = 	 {2016},
  editor = 	 {Balcan, Maria Florina and Weinberger, Kilian Q.},
  volume = 	 {48},
  series = 	 {Proceedings of Machine Learning Research},
  address = 	 {New York, New York, USA},
  month = 	 {20--22 Jun},
  publisher =    {PMLR}
}

@misc{sun2019rotateknowledgegraphembedding,
      title={RotatE: Knowledge Graph Embedding by Relational Rotation in Complex Space}, 
      author={Zhiqing Sun and Zhi-Hong Deng and Jian-Yun Nie and Jian Tang},
      year={2019},
      eprint={1902.10197},
      archivePrefix={arXiv},
      primaryClass={cs.LG},
}

@INPROCEEDINGS{10191314,
  author={Zhang, Yichi and Chen, Mingyang and Zhang, Wen},
  booktitle={2023 International Joint Conference on Neural Networks (IJCNN)}, 
  title={Modality-Aware Negative Sampling for Multi-modal Knowledge Graph Embedding}, 
  year={2023},
  volume={},
  number={},
  pages={1-8}}

@inproceedings{mousselly2018multimodal,
  title={A multimodal translation-based approach for knowledge graph representation learning},
  author={Mousselly-Sergieh, Hatem and Botschen, Teresa and Gurevych, Iryna and Roth, Stefan},
  booktitle={Proceedings of the seventh joint conference on lexical and computational semantics},
  pages={225--234},
  year={2018}
}

@INPROCEEDINGS{8852079,
  author={Wang, Zikang and Li, Linjing and Li, Qiudan and Zeng, Daniel},
  booktitle={2019 International Joint Conference on Neural Networks (IJCNN)}, 
  title={Multimodal Data Enhanced Representation Learning for Knowledge Graphs}, 
  year={2019},
  volume={},
  number={},
  pages={1-8}}

@article{lu2022mmkrl,
  title={MMKRL: A robust embedding approach for multi-modal knowledge graph representation learning},
  author={Lu, Xinyu and Wang, Lifang and Jiang, Zejun and He, Shichang and Liu, Shizhong},
  journal={Applied Intelligence},
  volume={52},
  number={7},
  pages={7480--7497},
  year={2022},
  publisher={Springer}
}

@inproceedings{wang2021visual,
  title={Is visual context really helpful for knowledge graph? A representation learning perspective},
  author={Wang, Meng and Wang, Sen and Yang, Han and Zhang, Zheng and Chen, Xi and Qi, Guilin},
  booktitle={Proceedings of the 29th ACM international conference on multimedia},
  pages={2735--2743},
  year={2021}
}

@article{zhang2022knowledge,
  title={Knowledge graph completion with pre-trained multimodal transformer and twins negative sampling},
  author={Zhang, Yichi and Zhang, Wen},
  journal={arXiv preprint arXiv:2209.07084},
  year={2022}
}

@inproceedings{wang2023tiva,
  title={TIVA-KG: A multimodal knowledge graph with text, image, video and audio},
  author={Wang, Xin and Meng, Benyuan and Chen, Hong and Meng, Yuan and Lv, Ke and Zhu, Wenwu},
  booktitle={Proceedings of the 31st ACM international conference on multimedia},
  pages={2391--2399},
  year={2023}
}

@article{song2024temporal,
  title={Temporal relevance for representing learning over temporal knowledge graphs},
  author={Song, Bowen and Amouzouvi, Kossi and Xu, Chengjin and Wang, Maocai and Lehmann, Jens and Vahdati, Sahar},
  journal={Semantic Web},
  volume={15},
  number={6},
  pages={2695--2711},
  year={2024},
  publisher={SAGE Publications Sage UK: London, England}
}

@article{wang2023survey,
  title={A survey on temporal knowledge graph completion: Taxonomy, progress, and prospects},
  author={Wang, Jiapu and Wang, Boyue and Qiu, Meikang and Pan, Shirui and Xiong, Bo and Liu, Heng and Luo, Linhao and Liu, Tengfei and Hu, Yongli and Yin, Baocai and others},
  journal={arXiv preprint arXiv:2308.02457},
  year={2023}
}

@ARTICLE{9961954,
  author={Zhu, Xiangru and Li, Zhixu and Wang, Xiaodan and Jiang, Xueyao and Sun, Penglei and Wang, Xuwu and Xiao, Yanghua and Yuan, Nicholas Jing},
  journal={IEEE Transactions on Knowledge and Data Engineering}, 
  title={Multi-Modal Knowledge Graph Construction and Application: A Survey}, 
  year={2024},
  volume={36},
  number={2},
  pages={715-735}}

@inproceedings{10.1145/3726302.3730082,
author = {Wang, Yunpeng and Ning, Bo and Wang, Xin and Liu, Chengfei and Li, Guanyu},
title = {Segmentation Similarity Enhanced Semantic Related Entity Fusion for Multi-modal Knowledge Graph Completion},
year = {2025},
isbn = {9798400715921},
publisher = {Association for Computing Machinery},
address = {New York, NY, USA},
booktitle = {Proceedings of the 48th International ACM SIGIR Conference on Research and Development in Information Retrieval},
pages = {1176–1185},
numpages = {10},
location = {Padua, Italy},
series = {SIGIR '25}
}

@inproceedings{
zhang2025multiple,
title={Multiple Heads are Better than One: Mixture of Modality Knowledge Experts for Entity Representation Learning},
author={Yichi Zhang and Zhuo Chen and Lingbing Guo and yajing Xu and Binbin Hu and Ziqi Liu and Wen Zhang and Huajun Chen},
booktitle={The Thirteenth International Conference on Learning Representations},
year={2025}
}

@ARTICLE{11192270,
  author={Shen, Jinqing and Xu, Chengjin and Liu, Yingqi and Jiang, Xuhui and Li, Jiaming and Huang, Zhenxin and Lehmann, Jens and Chen, Xuesong},
  journal={IEEE Access}, 
  title={Learning Temporal Knowledge Graphs via Time-Sensitive Graph Attention}, 
  year={2025},
  volume={13},
  number={},
  pages={178517-178526}}

@inproceedings{chen-etal-2025-noise,
    title = "Noise-powered Multi-modal Knowledge Graph Representation Framework",
    author = "Chen, Zhuo  and
      Fang, Yin  and
      Zhang, Yichi  and
      Guo, Lingbing  and
      Chen, Jiaoyan  and
      Pan, Jeff Z.  and
      Chen, Huajun  and
      Zhang, Wen",
    editor = "Rambow, Owen  and
      Wanner, Leo  and
      Apidianaki, Marianna  and
      Al-Khalifa, Hend  and
      Eugenio, Barbara Di  and
      Schockaert, Steven",
    booktitle = "Proceedings of the 31st International Conference on Computational Linguistics",
    month = jan,
    year = "2025",
    address = "Abu Dhabi, UAE",
    publisher = "Association for Computational Linguistics",
    pages = "141--155"
}

\appendix









\end{document}